\documentclass[sigconf]{acmart}

\AtBeginDocument{%
  \providecommand\BibTeX{{%
    \normalfont B\kern-0.5em{\scshape i\kern-0.25em b}\kern-0.8em\TeX}}}

\setcopyright{none}
\acmDOI{}
\renewcommand\footnotetextcopyrightpermission[1]{} 
\usepackage{cleveref}
\usepackage[normalem]{ulem}
\usepackage{colortbl}
\usepackage{booktabs}    
\usepackage{multirow}    
\usepackage{microtype} 
\usepackage{subfigure}
\usepackage{booktabs}

\makeatletter
\gdef\acmConference@shortname{}
\gdef\acmConference@name{}
\gdef\acmConference@date{}
\gdef\acmConference@venue{}
\gdef\@acmBooktitle{} 
\makeatother

\begin{document}

\title{When Images Speak Louder: Mitigating Language Bias-induced Hallucinations in VLMs through Cross-Modal Guidance}


\newcommand{\authorBlock}{%
\begin{tabular}{@{}c@{}}
Jinjin Cao\quad Zhiyang Chen$^{*}$\quad Zijun Wang\quad
Liyuan Ma\quad Weijian Luo\quad Guojun Qi$^{*}$\\[0.6ex]
MAPLE Lab, Westlake University\\
\texttt{\small caojinjin@westlake.edu.cn, chenzhiyang@westlake.edu.cn, guojunq@gmail.com}
\end{tabular}%
}
\author{\authorBlock}



\begin{abstract}

Vision-Language Models (VLMs) have shown solid ability for multimodal understanding of both visual and language contexts. However, existing VLMs often face severe challenges of hallucinations, meaning that VLMs tend to generate responses that are only fluent in the language but irrelevant to images in previous contexts. To address this issue, we analyze how language bias contributes to hallucinations and then introduce Cross-Modal Guidance(CMG), a training-free decoding method that addresses the hallucinations by leveraging the difference between the output distributions of the original model and the one with degraded visual-language attention. In practice, we adaptively mask the attention weight of the most influential image tokens in selected transformer layers to corrupt the visual-language perception as a concrete type of degradation. Such a degradation-induced decoding emphasizes the perception of visual contexts and therefore significantly reduces language bias without harming the ability of VLMs. 
In experiment sections, we conduct comprehensive studies. All results demonstrate the superior advantages of CMG with neither additional conditions nor training costs. We also quantitatively show CMG can improve different VLM's performance on hallucination-specific benchmarks and generalize effectively. 
\end{abstract}

\maketitle
\renewcommand{\shortauthors}{}
\pagestyle{plain}

\section{Introduction}
\label{sec:introduction}


%

\begin{figure}
\centering     
\subfigure[]{\label{fig:Intro_1}
                     \includegraphics[width=80mm]{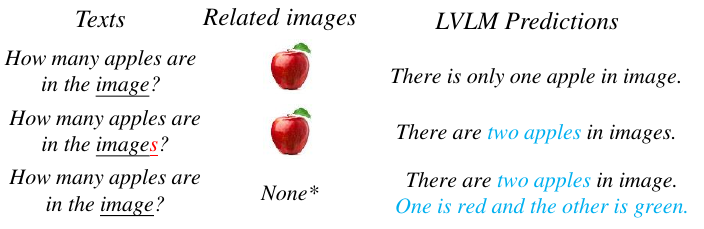}}
\subfigure[]{\label{fig:Intro_2}
                     \includegraphics[width=80mm]{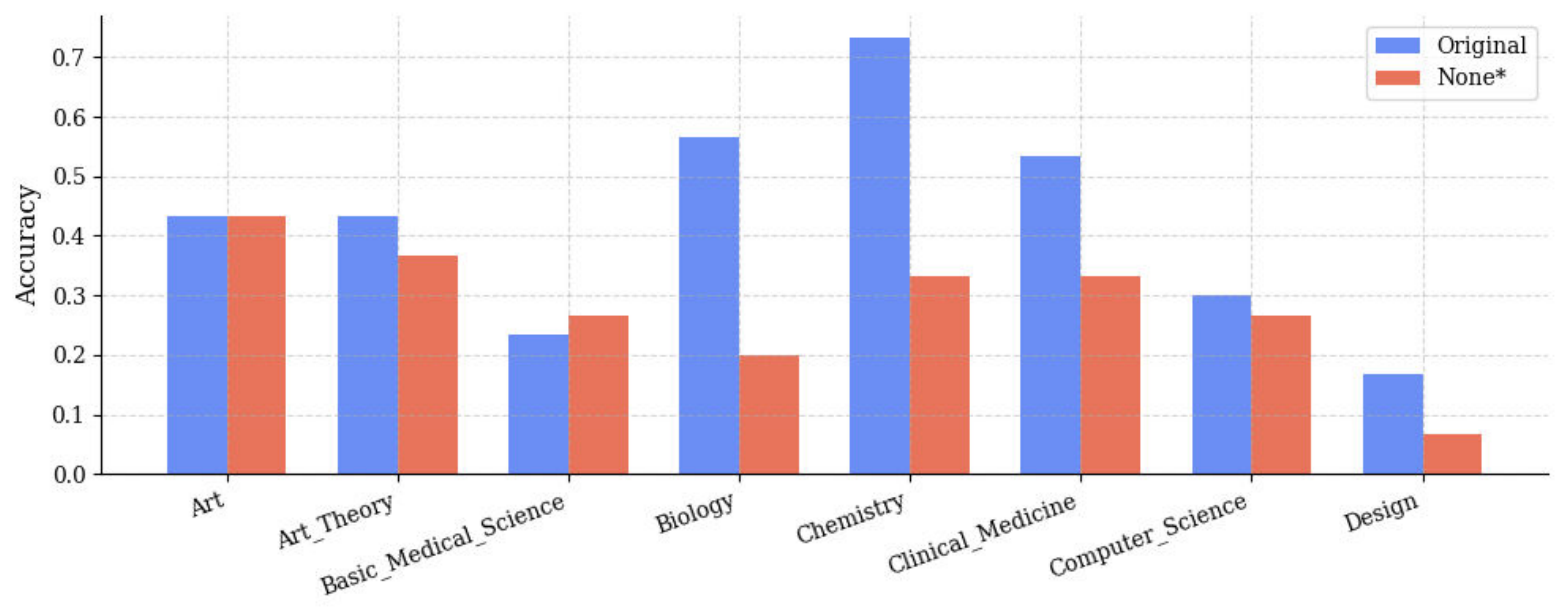}}
\caption{\textbf{An illustration of hallucinations induced by language bias}. (a) Examples of hallucinations induced by language bias in VLMs. The \textcolor[HTML]{00B0F0}{blue} words are hallucination contents. (b) Accuracy in MMMU\cite{yue2024mmmu} Benchmark on LLaVA-v1.5-7B. 'None*' represents images that are removed from the visual question input. }
\label{fig:Intro}
\end{figure}

Vision-Language Models (VLMs) like GPT-4o\cite{openai2023gpt4}, LLaVA-Series\cite{llama-adapter, llama-adapter-v2, llava}, QwenVL-Series\cite{Qwen-VL, Qwen2VL, qwen-lm}, and others \citep{chen2023sharegpt4v, pandagpt, palm-e, li2021align, chen2020uniter, cho2021unifying, wang2021simvlm, zhan2025griffon}, have shown solid abilities in multi-modal information perception and reasoning, sparking a new wave of applications of modern artificial intelligence. 
Despite powerful capacities, many recent researchers have found that VLMs sometimes suffer from hallucinations: \emph{VLMs often tend to generate incorrect responses that are irrelevant to image inputs in previous contexts}. For instance, as Figure \ref{fig:Intro_1} shows, when we ask the VLM to count the apples in the image, the answer seems to depend more on whether the world \emph{images} is singular or plural, rather than the factual visual content. Even though we provide no images, VLMs can still generate a plausible answer. This phenomenon may be denoted as language bias, in that the model generates responses following the learned language pattern and ignoring visual information.

\begin{figure*}
\centering     
\subfigure[]{\label{fig:method_1}
                     \includegraphics[width=80mm]{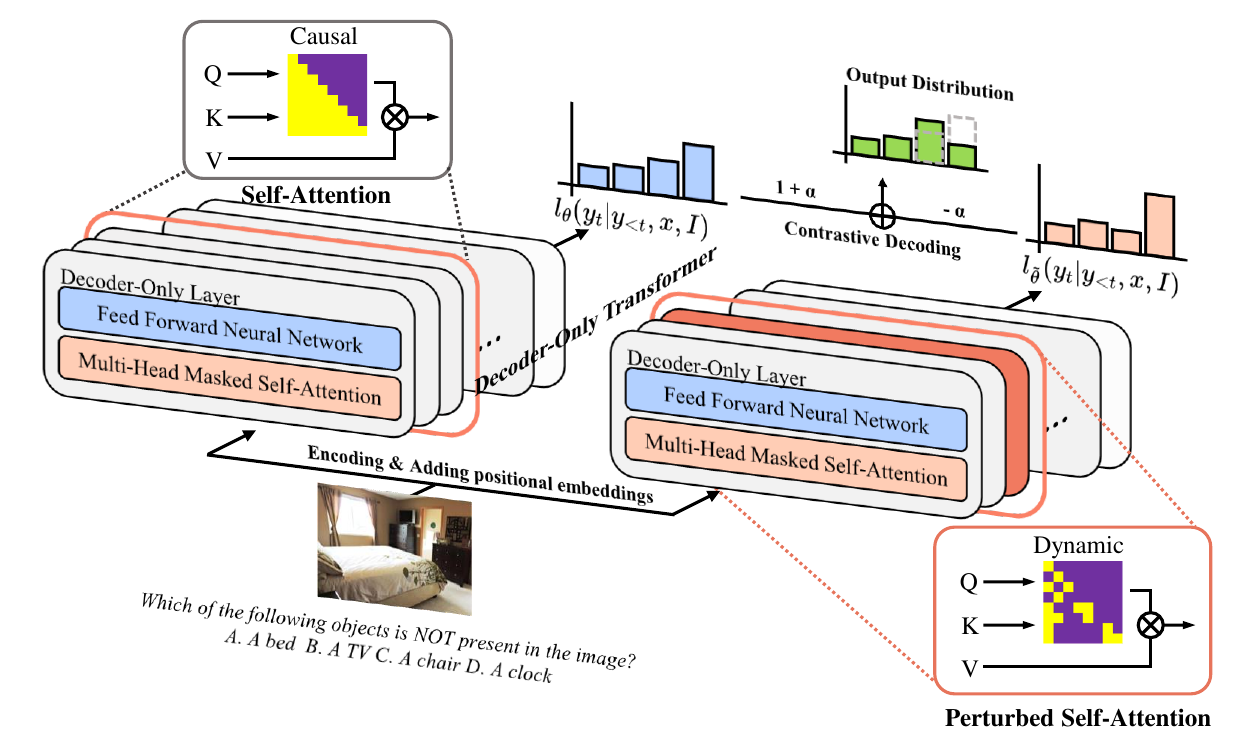}}
\subfigure[]{\label{fig:method_2}
                     \includegraphics[width=80mm]{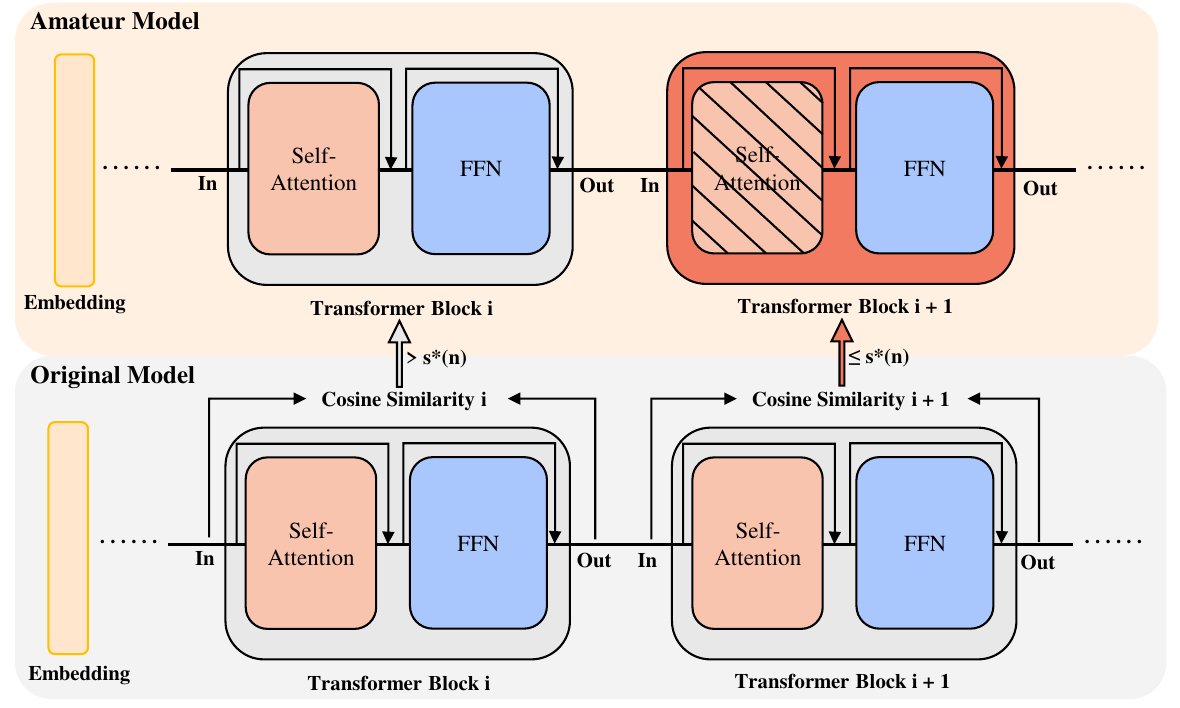}}
\caption{\textbf{Architecture of Cross-Modal Guidance.} CMG utilizes a perturbed self-attention map to amplify language priors in the underlying decoder-only transformer backbone. The original self-attention uses a causal mask, while the perturbed self-attention map replaces it with a dynamic mask, which varies from different samples. Perturbed self-attention is applied to several dynamically selected decoder-only layers. CMG contrasts the two distributions to correct hallucinations from the original outputs.}
\label{fig:method}
\end{figure*}

Such a property brings uncontrollable risks for users when using VLMs, preventing the broader use of VLMs. To mitigate this issue, some existing works have proposed various solutions, such as prompt-engineering-based methods \cite{gunjal2024detectingpreventinghallucinationslarge}, post-training using human feedback data \cite{ouyang2022traininglanguagemodelsfollow}, and developing different inference strategies \citep{leng2024mitigating,park2024convis}. 

In this paper, we focus on the inference of VLMs. We introduce \emph{Cross-Modal Guidance(CMG)}, a training-free inference algorithm for vision-language models that can significantly reduce hallucinations. CMG first corrupts the visual-language attention by randomly masking attention weights in certain transformer-based VLM decoder layers. Then CMG computes the inference logit values by adding a scaled difference term of the output logits using original and masked attention weights. Such an attention-corruption mechanism enhances the visual-language perceptions inside the neural networks, distinguishing CMG from previous methods such as VCD\citep{leng2024mitigating} that directly add Gaussian noises on input images. Besides, CMG is different from other previous methods such as ConVis\citep{park2024convis} that call expensive additional models to enhance the visual information. From a high level of view, the CMG \emph{make images speak louder} inside the neural network, therefore can address the insufficient visual perception of VLMs that potentially cause hallucinations.

In \cref{sec:Experiments}, we find that CMG can improve the generation performances of VLM with a significant effect of reducing hallucinations. 
CMG also outperforming its counterparts with VCD and ConVis in POPE and HallusionBench benchmark. 
On the MME benchmark, CMG surpass VCD by a large margin, reaching \textbf{13.54\%} performance gain and also exceeds ConVis by \textbf{+8.0}.
On the POPE benchmark, LLaVA-v1.5-7B with CMG achieves an overall accuracy value of \textbf{85.48}, outperforming its counterparts with VCD and ConVis. On HallusionBench, CMG exceeds VCD by \textbf{+7.1} and ConVis by \textbf{+6.3} in accuracy with no additional training, marking a leading performance among training-free inference approaches.

Our contributions are summarized as follows:
\begin{itemize}
    \item We introduce CMG, a training-free inference method that can effectively reduce models' hallucinations by enhancing the visual-language perceptions through random attention masks;
    \item We identify the insufficient visual-attention connections as one of the causes of hallucinations with rigorous evidence;
    \item We quantitatively evaluate CMG and show its solid performances on multiple benchmarks. 
\end{itemize}

\section{Related Work}
\label{sec:Related Work}
\subsection{Hallucinations in Vision-Language Models}
Vision-Language Models(VLMs)\cite{minigpt-4, openai2023gpt4, Qwen-VL, llava, chen2023sharegpt4v, maaz2023videochatgpt, zhang2023videollama, gong2023multimodalgpt} have revolutionized based on the development of Large Language Models(LLMs). VLMs can receive both visual and textual input, generating text responses iteratively. Specifically, to process image inputs, VLMs use an image encoder and linear projections to align text and image embeddings; for instance, LLaVA-v1.5\cite{llava} uses CLIP\cite{clip} as its image encoder. However, despite the powerful ablities, misalignment emerges in VLMs. Hallucination generally refers to cases where generated responses include information unrelated to image content. Some benchmarks\cite{Li-pope, mme, chen2023mitigating} are collected to evaluate hallucinations. 

\subsection{Content-Aware Decoding}

Proper decoding (inference) methods are essential for both large language models and vision-language models to get optimal performances. From a high level of view, decoding methods can generally be categorized into search and sampling algorithms \cite{li2023contrastivedecodingopenendedtext}. 

Search methods, like greedy and beam search, produce accurate results but often lead to tedious and repetitive outputs. In contrast, sampling methods, such as nucleus sampling \cite{holtzman2020curiouscaseneuraltext}, generate more diverse text but can suffer from unnatural topic shifts. To address these issues, content-aware decoding\cite{li2023contrastivedecodingopenendedtext, shi2023trustingevidencehallucinatecontextaware} was proposed for large language models, leveraging the difference between two output probabilities to construct a new and potentially enhanced output distribution. Similar ideas arose in the literature of vision-language models in recent years. VCD\cite{leng2024mitigating} contrasts distribution with original and distorted image inputs to reduce statistic bias and language priors in LVMs. Also focusing on image input, ConVis\cite{park2024convis} utilizes an additional text-to-image model to regenerate the caption of the original image, and then uses the difference in details between the new image and the original image to guide the generation. However, these methods only distort image input, leaving in-depth research on the black-box nature of VLMs. In this paper, we focus on the transformer attention mechanism, which has not been studied in previous research. 
We found language bias induces the emergence of hallucinations. By destroying the modal attention connection between image and text to contrast with the original distribution, we strengthen the reliance on visual context and eliminate the influence of language bias. 
\section{The Proposed Method}
\label{sec:Method}

\begin{figure}[htpb]
    \centering
    \includegraphics[width=1.0\linewidth]{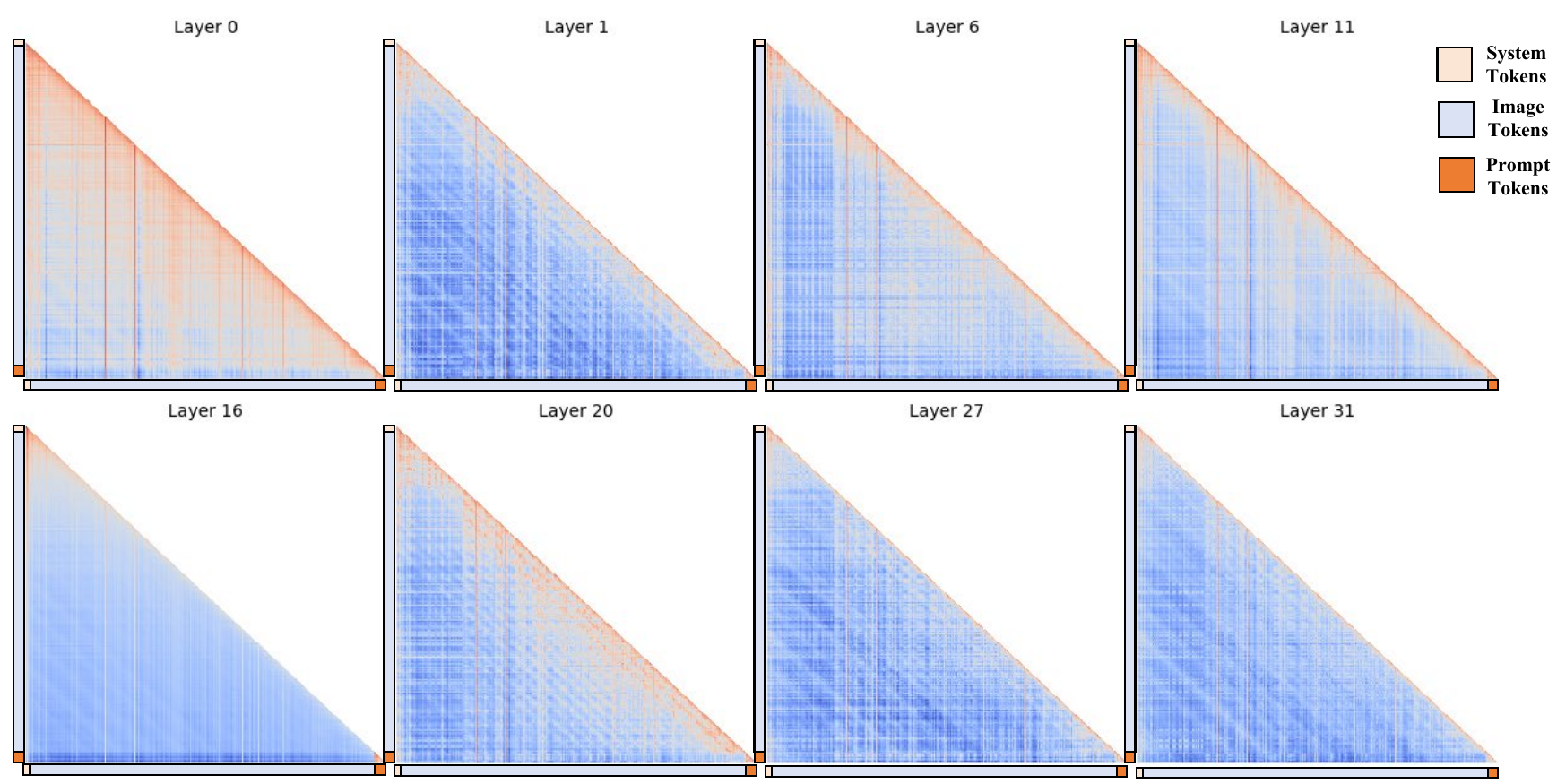}
    \caption{\textbf{Visualization of Attention Weights Changes Across Transformer Layers.} The overall trend of image token weight ratio is getting lower as the number of transformer layers increases.}
    \label{fig:method_3}
\end{figure}

\subsection{Preliminaries}
A Vision-Language Model(VLM) parameterized by $\theta$ with an autoregressive, autoencoding or encoder-decoder architecture pretrained on a large corpus of millions to trillions of tokens. VLMs are usually adapted for a specific task, for example, image captioning or vision-question answering, by fine-tuning in a relatively small dataset compared to pretraining. 

VLMs receive interleaved images(or videos) and texts as input, generates coherent and fluent texts as answers. 
Specifically, we consider text input of length $n$, denoted as $x = \{x_1, x_2, ..., x_n\}$  and visual input of length $m$, denoted as $I = \{I_1, I_2, ..., I_m\}$. After decoding, we acquire a text sequence of length $k$ denoted as $y = \{y_1, y_2, ..., y_k\}$. 

The text output $y$ is generated auto-regressively by the underlying language model $p_{\theta}$. During decoding, tokens are generated iteratively, each conditioned on the preceding context: 
\begin{equation}
  p_\theta(y|x, I) = \prod_{t=1}^k {p_\theta(y_t|y_{<t}, x, I)}
\end{equation}
where $p_\theta(y_t|y_{<t}, x, I)$ denotes the next token distribution. We use different subscripts to denote different weights of language model: $p_{\theta}$ is the original VLM, $\tilde{p}_{\theta}$ is the amateur VLM where the model result is less accurate than the original.

In detail, the relationship between the output distribution and the direct output of logits is:
\begin{equation}
    p_\theta(y|x, I) = \text{softmax}[l_\theta(y|x, I)],
\end{equation}
where $l_\theta(y|x, I)$ denotes as the logits output of language model.

\subsection{Language Bias Raises Hallucinations in VLMs}


Language bias refers to answers being strongly biased towards textual part of input questions while the importance of visual part is overlooked. This bias strongly influences the responses of VLMs, leading to a preference for content closely related to the language pretraining data, while being weakly or even completely unrelated to the current visual input. In \cref{fig:Intro_1}, we show two typical cases of hallucination in VLMs. By simply replacing ``image" in the question with ``images", VLM outputs two completely different answers under the same image input conditions. The other case is VLM responds to text prompt when there is no image input at all. In \cref{fig:Intro_2}, compared to the baseline, when completely deleting image input, the performance on the MMMU Benchmark does not degenerate to completely inaccurate. Different degrading scores at subsets implies that samples are affected by language bias in separate degrees. The scores on some subtasks like Art are greater than random selection probability, which indicates that the baseline's answers are affected by inherent language bias.


\begin{figure}
\centering     
\subfigure[]{\label{fig:language_1}
                     \includegraphics[width=80mm]{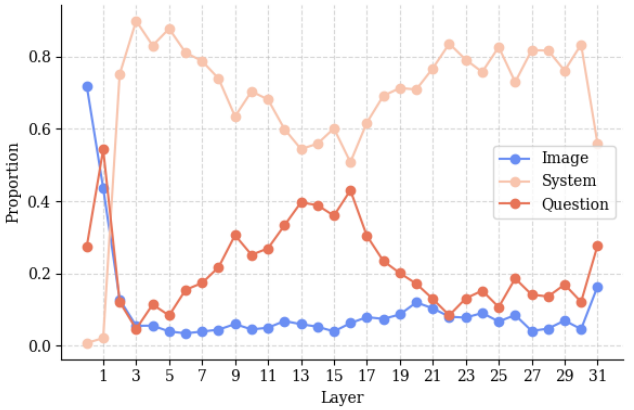}}
\subfigure[]{\label{fig:language_2}
                     \includegraphics[width=80mm]{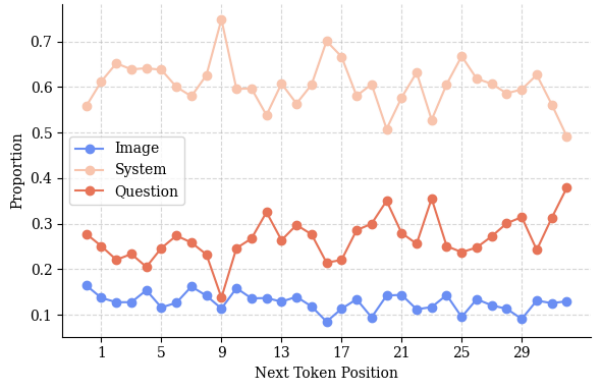}}
\caption{\textbf{Variation in Attention Weight Proportions Across Token Sequence Parts.} (a) The proportion of image attention weights changes with transformer layer. (b) The proportion of image attention weights changes with generated token sequence lengthens.}
\label{fig:language}
\end{figure}

However, the image tokens usually outnumber the text tokens in input sequence of VLMs. Taking LLaVA-v1.5-7B as an example, the image tokens are typically encoded into a sequence of 576 tokens, while the text token length is about one-tenth of the image length. So why the impact of language bias on the output can sometimes outweigh the influence of the images? As shown in \cref{fig:method_3}, we discover image attention weights drop sharply in shallow layers, maintaining low weights in deeper layers. \cref{fig:language_1} clearly shows the image attention weights decay sharply in the first few shallow layers, only rising sightly in the last transformer layers. In contrast, in the shallow layer, the ratio of text attention weights increases significantly, especially the ratio of system tokens even exceeds that of question tokens containing key information to answers. This implies the role of image token is largely overlooked than text tokens as transformer layer goes deeper, which induces hallucination in VLMs.


We further investigate the change in attention weights  as the generated token sequence grows. As shown in \cref{fig:language_2}, the proportion of image attention weights also decreases gradually, while text tokens maintain a high proportion. This phenomenon can explain why hallucinations are more likely to occur when generating long contexts. 

Findings in language bias in VLMs highlight that language bias contributes to hallucinations in VLMs, and thus we ought to mitigate it by enhancing model's attention on images. 

\subsection{Cross-Modal Guidance}

\subsubsection{Constructing Amateur Model with Attention Mask}
\begin{figure}[htpb]
    \centering
    \includegraphics[width=60mm]{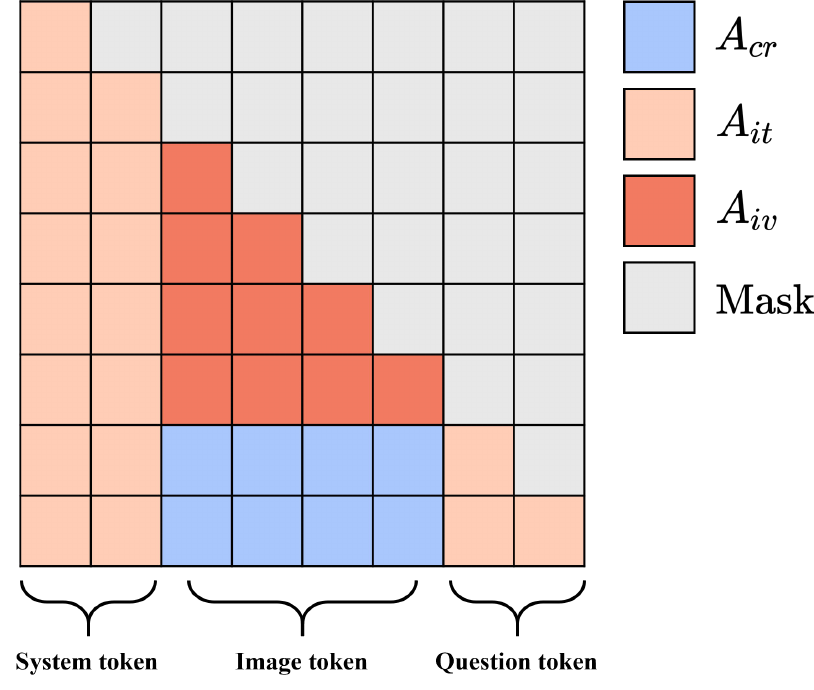}
    \caption{\textbf{Self Attention Weights with Causal Mask}}
    \label{fig:method_0}
\end{figure}

As shown in \cref{fig:method_0}, the self attention $A$ in transformer blocks consists of three types: inter-visual attention $A_{iv}$, inter-textual attention $A_{it}$, and cross-modal attention $A_{cr}$. 
\begin{equation}
    A = A_{iv} \cup A_{it} \cup A_{cr}
\end{equation}
As we discussed above, in order to mitigate language bias, we need to enhance both inter-visual attention and cross-modal attention to make better use of visual contents.

If we retain only inter-textual attention $A_{it}$, VLMs degrade to a model that is forced to generate the output distributions solely biased towards language questions.
By comparing this biased output distributions with original ones, we form the pointwise mutual information (PMI) between the final output $y$ and visual-related attentions $A_{cr}$ and $A_{iv}$ similar to those in CFG and contrastive decoding \cite{shi2023trustingevidencehallucinatecontextaware}, and it can be used to adjust the original output distributions of VLMs,
\begin{equation}
    \tilde p_\theta(y|x, I) \propto p_\theta(y|x, I) \left(
        \frac{
              p_\theta(y|x, I;  A_{cr}, A_{iv}, A_{it})
        }{
              p_{\theta}(y|x, I;  \emptyset, \emptyset, A_{it})
        }
    \right)^\alpha
\label{equ:first}
\end{equation}
where $p_{\theta}(y|x, I; \emptyset, \emptyset, A_{it})$ denotes both cross-modal attention and inter-visual attention are  completely masked out. This yields a preference over an output $y$ that is more likely to be generated with these visual-related attentions rather than without them. 

However, as shown in \cref{tab:MME_noIMG}, we find simply removing all visual-related attentions results in poor $p_{\widetilde{\theta}}$ that fails to generate correct answers in VLMs. We hypothesize that it would cause the collapse of the underlying VLM network as its attention structure was over-disturbed. Usually, a constraint shall be imposed on how much disturbance ought to be allowed.
This inspires us to only remove part of cross-modal and inter-visual attention weights using masks, and we set a maximum size for these attention masks to control how many attention weights can be removed. 



By removing part of cross-modal and inter-visual attention, we strengthen the role of inter-textual attention that would enhance the language bias in VLMs. In this case, if the introduction of cross-modal and inter-visual attention leads to an increase in the probability of a responding word, we may deduce that this word should be highly related to the visual content. Thus we should favor these words during sampling, which can be achieved by adjusting the output distribution with a similar PMI ratio as in Eq.~\ref{equ:first}.


Formally, we adjust the original VLM's output distribution \\
$p_\theta(y|x, I; A_{cr}, A_{iv}, A_{it})$ to obtain a new one $\tilde p_\theta(y|x, I)$ as
\begin{align}
    &\ \tilde p_\theta(y|x, I) \propto q_\theta(y)    \left(
        \frac{
              q_\theta(y)
        }{
              q_{\theta}(y; M)
        }
    \right)^\alpha \label{equ.ratio}
\end{align}
where $M$ denotes the mask imposed on the attention map, that is
\begin{align}
    &\ M := M_{cr} \cup M_{iv}  \\
    &\ q_\theta(y) := p_\theta(y|x, I; A_{cr}, A_{iv}, A_{it}) \\
    &\ q_{\theta}(y; M) := \tilde{p}_{\theta}(y|x, I; A_{cr}\odot M_{cr}, A_{iv}\odot M_{iv}, A_{it} ) 
\end{align}
Here $M_{cr}$ and $M_{iv}$ are masks on cross-modal attention and inter-visual attention respectively, and we denote the masked VLM model by \emph{Amateur Model}.
These masks are applied to self-attentions in Eq.~\ref{equ.sa} below
\begin{align}
    SA(Q, K, V; M) &= \text{Softmax}(\frac{QK^T}{\sqrt{d}} \odot M)V \\
                   &= \text{Softmax}(A \odot M)V.
    \label{equ.sa}
\end{align}

\subsubsection{Dynamically Masking Amateur Models}

Finding an optimal mask $M$ subject to a maximum size $n_0$ can be formulated by maximizing the divergence between $q_\theta(y)$ and 
$q_{\theta}(y; M)$,
\begin{align}
    \max_{M} \ &\text{KL}[ q_{\theta}(y; M), q_\theta(y)] \\
    \text{s.t.} \quad&\|1-M_{cr}\|_0 + \|1-M_{iv}\|_0 \leq n_0 \label{eqn:optim_st}\\
    &M_{cr}\cup M_{iv} \in \{0,1\}^N
\end{align}
where $\text{KL}$ is the KL divergence, $\|\cdot\|_0$ is the $\ell_0$-norm that accounts the number of non-zero elements, and $N$ is the total number of candidate positions to mask in a VLM model. By maximizing the divergence, we will obtain a masked model that maximizes the contrast with the original model. In this way, the ratio of $q_\theta (y)$ and $q_\theta(y;M)$ is more likely to increase \footnote{We cannot directly maximize the ratio of $q_\theta (y)$ and $q_\theta(y;M)$ in Eq.~\ref{equ.ratio} since the groundtruth output $y$ is unknown beforehand in the inference.} if the likelihood of generating an output $y$ is more likely after the masked inter-visual and cross-modal attentions are filled back to the model. This strengthens the role of these visual-related attentions, which could mitigate the language bias often related with text-only attentions.

Unfortunately, directly optimizing the above constrained objective is intractable as it is a NP-hard problem. We provide two dynamic strategies to determine which attention weights to mask in some selected layers. The basic idea is to find the part of attention weights making the largest contribution to the output of the original VLM model, by removing which its output distribution $q_\theta(y)$ could be greatly changed.  

\paragraph{Dynamic attention masking}
\begin{figure}
    \centering
    \includegraphics[width=80mm]{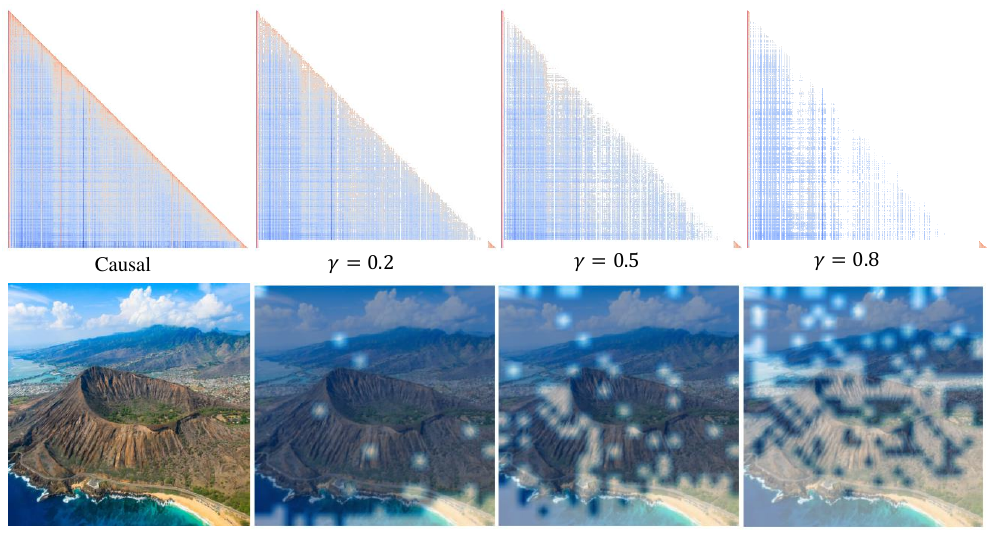}
    \caption{Inter-visual attention masks with different $\gamma$. The masks are visualized by an average of mask values associated with a pixel patch. The asked question for this example in the VLM model is \emph{Describe this photo in detail}. }
    \label{fig:method_mask}
\end{figure}

\begin{figure}[htpb]
    \centering
    \includegraphics[width=80mm]{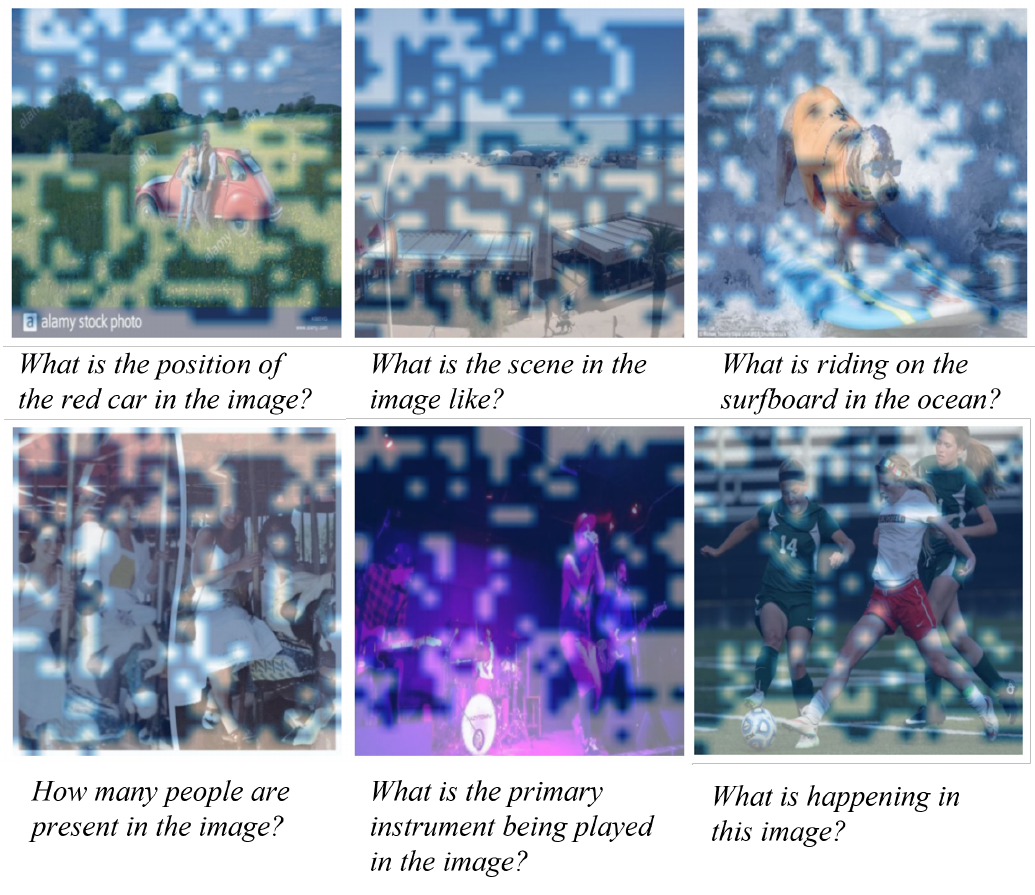}
    \caption{\textbf{Attention Mask When $\gamma$=0.5.} The whiter part is the masked part.}
    \label{fig:appendix_1}
\end{figure}

Attention weights determine the level of importance each element contributes to the model's output. We partially mask the largest $\gamma$-portion of attention weights in $A_{iv}$ and $A_{cr}$, resulting in
\begin{equation}
    \widetilde{SA}(Q, K, V; M) = \text{Softmax}(A \odot M(\gamma))V
    \label{equ.7}
\end{equation}
\begin{equation}
    M(\gamma) = A_{cr}(\gamma) \cup A_{iv}(\gamma)
    \label{equ.7_1}.
\end{equation}

As shown in \cref{fig:method_mask}, masking positions are usually related to objects in an image, which would make the answer to the asked question of ``Desribe this photo in detail" more related to the relevant visual information in the image.
As  $\gamma$ increases, the masks would become more selective in retaining visual parts in answering the question. \cref{fig:appendix_1} provides more examples to illustrate the relationship between the masking position and the question.

\begin{figure}
    \centering
    \includegraphics[width=80mm]{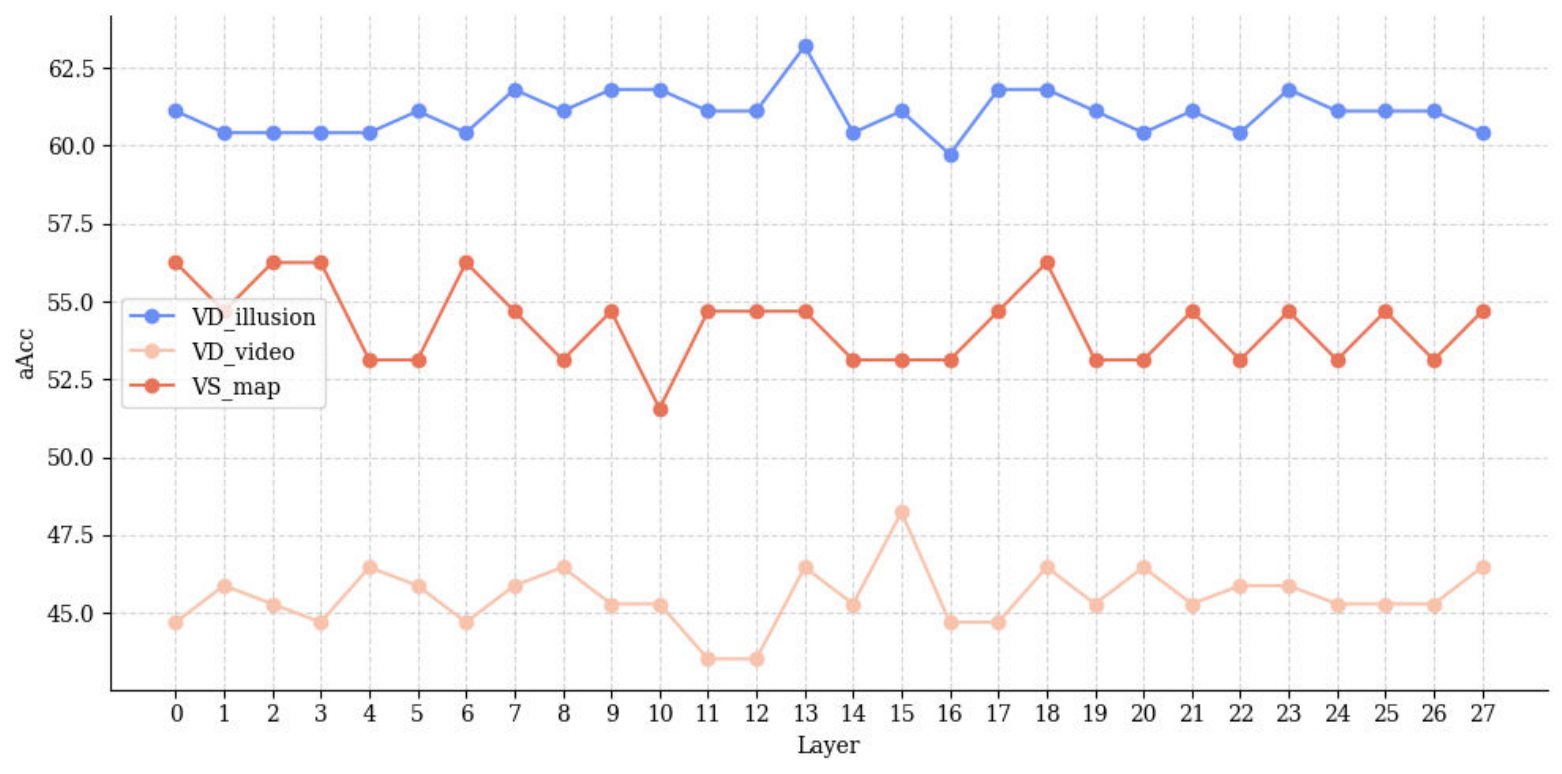}
    \caption{All accuracy score(aAcc) in HallusionBench when masking different transformer layers in Qwen-2-VL-2B-Instruct\cite{Qwen2VL}. Each time only one layer is masked.}
    \label{fig:Layer_1}
\end{figure}

\paragraph{Dynamic layer selection}
We also observe that different transformer blocks contribute unequally to the output distribution. As shown in \cref{fig:Layer_1}, when only a single layer is selected to apply the above dynamic attention masking approach, we find that the optimal layer index varies across subsets in terms of accuracy scores. 


This suggests that different layers have different effects on the result. We must carefully decide which layers to mask the attention weights by adopting a dynamic layer selection strategy. Formally, we determine which layers need to be selected by calculating the cosine similarities between the layer input $X$ and output distribution $Y$. We only choose those layers whose cosine similarity is sufficiently small, i.e., the layer output changes a lot from its input. In other word, if a layer changes its inputs a lot to give its outputs, it is considered playing an important role in the model. So selecting to mask its attentions could more significantly disturb the original VLM model. 

Formally, given the number of $n$ layers, the dynamic layer selection can be defined as: 
\begin{equation}
    s(i) = cos(X_i, Y_i) = \frac{X_i \cdot Y_i}{\left\| X_i \right\|_2\left\| Y_i \right\|_2}
    \label{equ.8}
\end{equation}
\begin{equation}
    s^* = \mathop{AscentSort} (\{s(i)|i=1,\cdots,n\})[\tau \cdot n]
    \label{equ.9}
\end{equation}
\begin{equation}
    \mathbb{Z} = \{i | s(i) \leq s^*\}
    \label{equ.10}
\end{equation}
where $\tau$ denotes the proportion of layers that shall be selected, $s*$ is the cosine similarity at the smallest $\tau$ percentile,  and $\mathbb{Z}$ denotes the index set of selected layers.

After a layer is selected, the dynamic attention masking is applied to this layer. \cref{fig:method} shows the full Cross-Modal Guidance method to construct an amateur model. 



\textbf{Why we choose cosine similarity to determine layer importance?} Cosine similarity is a commonly used way to measure similarity between two vectors. Dot product and Euclidean distance also can be used to measure similarity between the output and input of a layer. \cite{liu2023dejavucontextualsparsity, chen2024streamliningredundantlayerscompress} suggest that the magnitude of hidden states in transformers tend to grow as the layer becomes deeper. The dot product and Euclidean distance are both influenced by the vector magnitude, which means that they also changes with the layer. Consequently, we adopt cosine similarity, which only depends on the vector direction.

\section{Experiments}
\label{sec:Experiments}
\subsection{Experimental Setup}
\paragraph{Benchmarks.} To prove effectiveness of our method on mitigating language bias in LVLMs, we conduct experiments on three benchmarks. They are: 

\begin{figure}[htpb]
    \centering
    \includegraphics[width=1.0\linewidth]{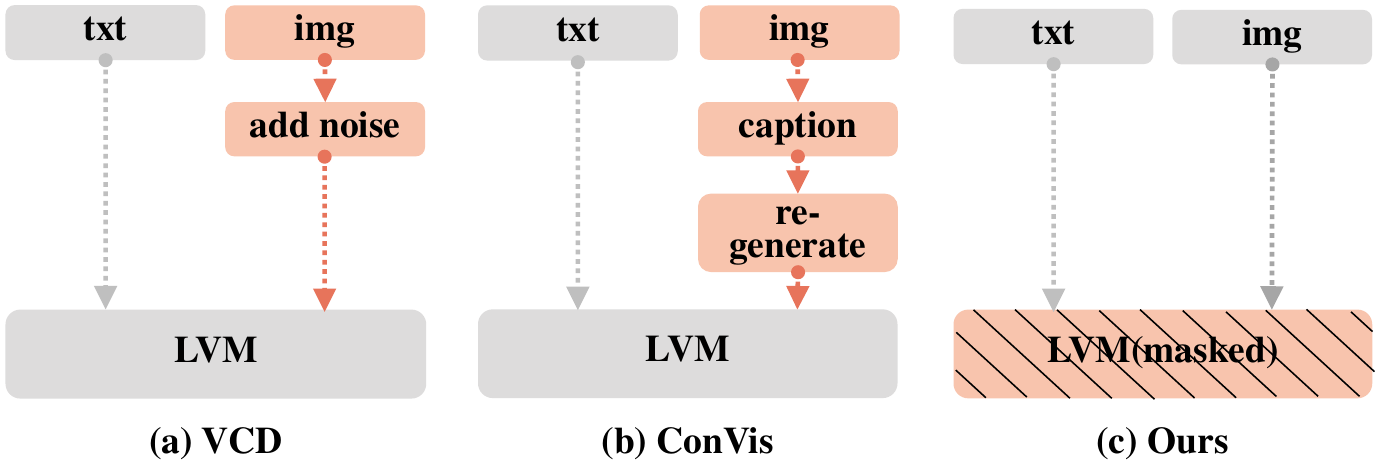}
    \caption{Comparison chart of VCD, ConVis and Ours}
    \label{fig:intro_3}
\end{figure}

\begin{itemize}
    \item Hallucination-related: HallusionBench\cite{guan2023hallusionbench}, POPE\cite{Li-pope}
    \item Comprehensive: MME\cite{mme} 
\end{itemize}

\paragraph{Compared Methods.} Current researches have designed various methods to construct an amateur model. As ~\cref{fig:Intro_2} shows, VCD\cite{leng2024mitigating} adds Gaussian noise to original images as the amateur's visual input. ConVis\cite{park2024convis} captions the original image 
transforms the original image input to caption prompts, and utilize text-to-image model to generate a new image based on captions, and utilize the difference between the original and re-generated image input. However, in these methods only the input of the model is focused, without in-depth research on the model decoding mechanism. The amateur input obtained by distorting the visual input not only changes the entire image distribution, but also becomes uncontrollable in the deep network of the model. Our method use the attention mask to dynamically corrupt amateur model's performance, fully considering features of different types of samples.

\paragraph{Inplementation Details.} We employ LLaVA-v1.5-7B \cite{llava}, Instructblip-7B\cite{dai2023instructblipgeneralpurposevisionlanguagemodels}, Qwen2-VL-7b\cite{Qwen2VL} and InternVL2.5-8b\cite{luo2024monointernvlpushingboundariesmonolithic} as our backbone model, using publicly available checkpoint weights. We set the top-p parameter to 0.9, beam search parameter to 5, temperature to 0.7 in our baseline. We set $\alpha$=0.3, $\gamma$=0.5, $\tau$=0.5 for hallucination-specific benchmarks, and $\alpha$=0.1 $\gamma$=0.5 $\tau$=0.1 for general benchmark MME. For both VCD and ConVis, the parameters employed are consistent with the optimal parameters provided in their respective papers.

\subsection{Results}

\begin{table*}
\centering
\caption{ Evaluation results on datasets designed for hallucinations}
\scalebox{0.9} {
\begin{tabular}{*{2}{c} c c c | c c c c}
\toprule
\multirow{2}{*}{Model} & \multirow{2}{*}{Method} & 
\multicolumn{3}{c|}{HallusionBench} &  \multicolumn{4}{c}{POPE} \\
\cmidrule(l){3-9} 
&& Question Pair Acc(qAcc) & Figure Acc(fAcc) & All Acc(aAcc) & Recall & Accuracy & Precision & Overall \\
\midrule
\multirow{4}{*}{\shortstack{LLaVA-v1.5 \\ 7b}} 
                               & Baseline         & 11.4 & 16.2 & 46.1 & 78.9 & 85.9 & 91.8 & 84.9 \\
                               & VCD        & 11.6 & \textbf{16.8} & 45.8 & 78.7 & 85.7 & 91.5 & 84.6 \\
                               & ConVis        & 11.2 & 15.9 & 45.3 & 78.8 & 85.9 & \textbf{91.8} & 84.8 \\
                               \cmidrule(l){2-9}
                               & Ours    & \textbf{11.9} & 16.5 & \textbf{46.2} & \textbf{79.3} & \textbf{86.3} & 91.7 & \textbf{85.4} \\
\midrule                               
\multirow{4}{*}{\shortstack{InstructBlip \\ 7b} }
                               & Baseline         & 17.1 & 20.9 & 51.1 & 77.1 & 85.2 & \textbf{92.1} & 83.9 \\
                               & VCD        & 17.1 & 20.9 & 51.0 & 76.67 & 84.93 & 91.98 & 83.63 \\
                               & ConVis        & 19.1 & 22.6 & 53.7 & 82.5 & 85.4 & 87.6 & 85.0 \\
                               \cmidrule(l){2-9}
                               & Ours    & \textbf{19.4} & \textbf{23.7} & \textbf{54.0} & \textbf{82.9} & \textbf{86.1} & 88.6 &\textbf{ 85.7} \\
\midrule
\multirow{4}{*}{\shortstack{Qwen2-VL \\ 7b}}  
                               & Baseline         & 43.7 & 39.0 & 68.5 & 78.3 & 87.4 & 95.7 & 86.1 \\
                               & VCD        & 40.7 & 34.7 & 65.6 & \textbf{81.7} & \textbf{88.6} & 94.7 & \textbf{87.7} \\
                               & ConVis        & 43.5 & 37.9 & 67.8 & 78.5 & 87.5 & \textbf{95.7} & 86.2 \\
                               \cmidrule(l){2-9}
                               & Ours    & \textbf{45.5} & \textbf{39.0} &\textbf{ 68.5} & 80.2 & 88.2 & 95.4 & 87.1 \\
\midrule
\multirow{4}{*}{\shortstack{InternVL2.5 \\ 8b}}  
                               & Baseline         & 41.1 & \textbf{43.1} & 68.1 & 84.2 & 89.6 & \textbf{94.3} & 89.0 \\
                               & VCD        & 39.3 & 37.6 & 65.2 & 83.5 & 88.7 & 93.1 & 88.1 \\
                               & ConVis        & 39.3 & 38.0 & 65.4 & 83.1 & 88.6 & 93.4 & 88.0 \\
                               \cmidrule(l){2-9}
                               & Ours    & \textbf{41.1} & 36.7 & \textbf{68.8} & \textbf{85.3} & \textbf{89.7} & 93.7 & \textbf{89.3} \\
\bottomrule
\end{tabular}
}

\label{tab:POPE&HallusionBench_official}
\end{table*}

\begin{table*}
\centering
\caption{Evaluation results on the MME benchmark.}
\scalebox{0.9} {
\begin{tabular}{*{12}{c}|c}
\toprule
\multirow{2}{*}{Model} & \multirow{2}{*}{Method} & 
\multicolumn{10}{c|}{Perception} & \multirow{2}{*}{Total}  \\
\cmidrule(l){3-12} 
&& existence & count & position & color & ocr  & poster & celebrity & scene & landmark & artwork \\
\midrule
\multirow{4}{*}{\shortstack{LLaVA-v1.5 \\ 7b}} 
                               & Baseline         & 185.00 & 93.33 & 113.33 & 160.00 & 117.50 & 90.81 & 90.29 & 142.00 & 136.50 & 114.00 & 1242.78 \\
                               & VCD        & 195.00 & 153.33 & 116.67 & 160.00 & 140.00 & 137.76 & 133.24 & 156.75 & 157.00 & 118.75 & 1468.49   \\
                               & ConVis        & \textbf{195.00} & 158.30 & \textbf{133.30} & 155.00 & 132.50 & \textbf{143.20} & \textbf{139.70} & 153.80 & 155.30 & 121.50 & 1487.60   \\
                               \cmidrule(l){2-13}
                               & Ours     & 190.00 & \textbf{158.33} & 126.67 & \textbf{160.00} & \textbf{147.50} & 142.52 & 134.12 & \textbf{154.50} & \textbf{158.00} & \textbf{124.00} & \textbf{1495.63}   \\
\midrule                               
\multirow{4}{*}{\shortstack{InstructBlip \\ 7b} }
                               & Baseline         & 180 & 65.00 & 55.00 & 138.33 & 95.00 & 129.59 & 153.53 & 158.25 & 101.00 & 130.00 & 1205.70 \\
                               & VCD        & 180.00 & \textbf{70.00} & \textbf{60.00} & 143.33 & 95.00 & \textbf{134.01} & 158.24 & 143.5 & 104.75 & 128.00 & 1216.83   \\
                               & ConVis        & 180.00 & 65.00 & 55.00 & 143.33 & 95.00 & 128.57 & 153.83 & \textbf{159.75} & \textbf{114.50} & 130.00 & \textbf{1224.98}   \\
                               \cmidrule(l){2-13}
                               & Ours             & \textbf{180.00} & 60.00 & 53.33 & \textbf{143.33} & \textbf{95.00} & 126.87 & \textbf{161.47} & 158.00 & 99.50 & \textbf{132.25} & 1209.76  \\
\midrule
\multirow{4}{*}{\shortstack{Qwen2-VL \\ 7b}}  
                               & Baseline         & 185.00 & 128.33 & 161.67 & 178.33 & \textbf{177.50 }& 164.29 & 125.00 & 144.75 & 154.25 & 131.75 & 1550.86 \\
                               & VCD        & 195.00 & 138.33 & 163.33 & 185.00 & 125.00 & 182.99 & 136.77 & \textbf{168.25} & 171.50 & 139.75 & 1605.92   \\
                               & ConVis        & \textbf{195.00} & \textbf{160.00} & 160.00 & 180.00 & 155.00 & 182.65 & \textbf{151.18} & 162.75 & \textbf{185.75} & 148.25 & 1680.58  \\
                               \cmidrule(l){2-13}
                               & Ours              & 190.00 & 155.00 & \textbf{165.00} & \textbf{185.00} & 170.00 & \textbf{185.37} & 149.41 & 160.75 & 184.25 & \textbf{149.75} & \textbf{1694.54}   \\
\midrule
\multirow{4}{*}{\shortstack{InternVL2.5 \\ 8b}}  
                               & Baseline         & 200.00 & 175.00 & \textbf{170.00} & 183.33 & 177.50 & 162.93 & 138.24 & 153.00 & 172.00 & 157.00 & 1688.99 \\
                               & VCD         & 200.00 & 175.00 & 151.67 & 175.00 & 177.50 & 167.01 & 136.74 & 155.50 & 169.00 & 159.75 & 1666.89 \\
                               & ConVis        & \textbf{200.00} & \textbf{175.00} & 158.33 & 175.00 & \textbf{177.50} & 165.65 & 140.88 & 154.75 & 172.00 & \textbf{160.50} & 1679.61   \\
                               \cmidrule(l){2-13}
                               & Ours              & 195.00 & 160.00 & 165.00 & \textbf{185.00} & 170.00 & \textbf{185.37} & \textbf{148.53} & \textbf{160.75} & \textbf{185.00} & 147.50 & \textbf{1702.00}   \\
\bottomrule
\end{tabular}
}

\label{tab:MME_official}
\end{table*}

\begin{table}
\centering
\caption{Evaluation results on the POPE with Different Size of Models.}
\scalebox{0.85} {

\begin{tabular}{*{6}{c}|c}
\toprule 
Model & Size & Method & Recall & Accuracy & Precision & Overall \\
\midrule
\multirow{2}{*}{InstructBlip} & 
\multirow{2}{*}{13B} &
Baseline & 35.3 & 60.5 & \textbf{95.7} & 51.6 \\ 
 &  & Ours & \textbf{40.2} & \textbf{62.7} & 94.3 & \textbf{56.4} \\
\midrule 
\multirow{2}{*}{Internvl2.5} & 
\multirow{2}{*}{26B} &
Baseline & \textbf{88.1} & 90.6 & 92.6 & \textbf{90.3} \\ 
 &  & Ours & 85.7 & \textbf{90.6} & \textbf{94.9} & 90.1 \\
\midrule 
\multirow{2}{*}{Qwen2VL} & 
\multirow{2}{*}{2B} &
Baseline & 82.3 & 89.0 & \textbf{95.0} & 88.2 \\ 
 &  & Ours & \textbf{83.7} & \textbf{89.2} & 94.0 & \textbf{88.5} \\
\midrule 
\multirow{2}{*}{Qwen2.5VL} & 
\multirow{2}{*}{7B} &
Baseline & 77.5 & 87.6 & \textbf{97.0} & 86.2 \\ 
 &  & Ours & \textbf{80.3} & \textbf{88.5} & 96.0 & \textbf{87.4} \\
\bottomrule
\end{tabular}
}
\label{tab:POPE_Size&Version_official}
\end{table}

\begin{table}
\centering
\caption{Evaluation results on the HallusionBench with Different Size of Models.}
\scalebox{0.9} {
\begin{tabular}{*{6}{c}}
\toprule 
Model & Size & Method & qAcc & fAcc & aAcc \\
\midrule
\multirow{2}{*}{InstructBlip} & 
\multirow{2}{*}{13B} &
Baseline & 22.9 & 16.2 & 49.9  \\ 
 &  & Ours & \textbf{24.4} & \textbf{18.2} & \textbf{53.6}  \\
\midrule 
\multirow{2}{*}{Internvl2.5} & 
\multirow{2}{*}{26B} &
Baseline & 46.6 & \textbf{47.7} & 71.5  \\ 
 &  & Ours & \textbf{47.9} & 45.1 & \textbf{72.1} \\
\midrule 
\multirow{2}{*}{Qwen2VL} & 
\multirow{2}{*}{2B} &
Baseline & 32.5 & 28.9 & 60.7 \\ 
 &  & Ours & \textbf{33.4} & \textbf{30.1} & \textbf{61.4} \\
\midrule 
\multirow{2}{*}{Qwen2.5VL} & 
\multirow{2}{*}{7B} &
Baseline & 40.4 & 35.8 & 65.7 \\ 
 &  & Ours & \textbf{47.0} & \textbf{45.7} & \textbf{70.5} \\
\bottomrule
\end{tabular}
}

\label{tab:HallusionBench_Size_official}
\end{table}

\begin{table}[htpb]
    \centering
    \caption{Additive effect evaluation of methods}
    \scalebox{0.8}{
        \begin{tabular}{@{}c l c c c | c c c c@{}}
            \toprule
            \multicolumn{2}{c}{Method} & \multicolumn{3}{c|}{HallusionBench} & \multicolumn{4}{c}{POPE} \\
            \cmidrule(r){3-5} \cmidrule(l){6-9}
            & & fAcc & qAcc & aAcc & Recall & Accuracy & Precision & Overall \\
            \midrule
            A & Baseline & 43.7 & 39.0 & 68.5 & 78.3 & 87.4 & \textbf{95.7} & 86.1  \\
            B & Ours & \textbf{45.5} & \textbf{39.0} & \textbf{68.5} & \textbf{80.2} & \textbf{88.2} & 95.4 & \textbf{87.1} \\
            \midrule
            $\text{C}_1$ & A + VCD & 40.7& 34.7& 65.6& 81.7& 88.6& \textbf{94.7}& 87.7\\ 
            $\text{C}_2$ & B + VCD & \textbf{47.2}& \textbf{47.0}&  \textbf{68.1}& \textbf{82.4}& \textbf{88.8} & 94.5& \textbf{88.0}\\
            \midrule
            $\text{D}_1$ & A + ConVis & \textbf{43.5}& 37.9& \textbf{67.8}& 78.5& 87.5& \textbf{95.7}& 86.2\\ 
            $\text{D}_2$ & B + ConVis & 39.0& \textbf{45.3}&  67.4& \textbf{81.3}& \textbf{88.3}& 94.5& \textbf{87.4}\\ 
            \bottomrule
        \end{tabular}
    }  
    \label{tab:Fusion_official}
\end{table}

\paragraph{Results on POPE benchmark.} The POPE benchmark evaluates object hallucinations by prompting VLMs to answer "yes" or "no" to questions regarding the existence of objects.
The results of the POPE experiments, including recall, accuracy, precision, and overall.
As demonstrated in ~\cref{tab:POPE&HallusionBench_official}, while conventional methods struggle with architecture upgrades (ConVis and VCD both dropping Overall Score on InternVL-2.5 versus its 89.0 baseline), our approach shows positive scaling (89.0→89.3 Overall) across model generations. This upward-compatible performance confirms our method's effectiveness as a universal solution for visual-language alignment.
In contrast to VCD and ConVis, which perform well only on older versions, our method achieves state-of-the-art performance across multiple backbones.

\paragraph{Results on HallusionBench benchmark.} HallusionBench is specific on hallucination in VLMs. It divides vision-question pairs into two types. The visual dependent questions relies on heavily on provided images, while the visual supplement questions can be answered without images referenced. The 1129 questions consists of diverse topics and formats, with binary choices of yes or no.
HallucinationBench encompasses a broader range of hallucination types compared to the POPE benchmark, that is limited to assessing object hallucinations. In ~\cref{tab:POPE&HallusionBench_official}, our method outperforms all listed baseline in both figure accuracy and overall accuracy. Question pair accuracy calculates the proportion of correctly answered questions when the picture is missing, figure accuracy measures the proportion of correctly answered images within the dataset, while overall accuracy represents the percentage of correctly answered questions. Specifically, we achieve a significant improvement in figure accuracy, demonstrating our enhanced capability to mitigate hallucinations in image understanding. 

\paragraph{Results on MME benchmark.} MME is a comprehensive benchmark measuring both the perception and reasoning abilities of VLMs. It consists of 14 subtasks which are divided into perception and reasoning categories. In perception category, MME evaluates coarse-grained recognition, fine-grained recognition and ocr abilities. The score for each subtask is the sum of the accuracy and accuracy+, where the latter refers to the score based on each image where all questions need to be answered correctly. 

As shown in \cref{tab:MME_official}, our method surpasses other decoding approaches in perception-related subtasks, which reflects fine-grained image understanding abilities. The scores presented in the table are the sum of accuracy and accuracy+. The accuracy+ metric is calculated at the image level, in other words, when all questions related to a single image are answered correctly, the accuracy+ for that image equals 100.

Our total score in the perception domain outperforms baseline, exceeding VCD by +62.08 and ConVis by +7.30 on average. Notably, our method achieves the highest scores in the ``color," ``scene", ``landmark," subsets, where hallucinations caused by language bias are particularly prevalent. In other subsets, our method also demonstrates competitive performance. The results on the MME benchmark demonstrate that our model continues to exhibit outstanding performance on general tasks.

\paragraph{Results on Different Size of Models.} \cref{tab:POPE_Size&Version_official} and \cref{tab:HallusionBench_Size_official} both show our model achieves robust performance across varying model scales, including 2B, 7B, 13B, and 26B parameter configurations, highlighting its scalability and architectural adaptability.

\paragraph{Results on the Method's Cumulative Effects.}  \cref{tab:Fusion_official} shows our method can be combined with both VCD and ConVis, achieving superior performance compared to using VCD or ConVis individually. However, this synergistic effect is not consistently observed, likely due to divergent optimization objectives among the different methods.

\section{Discussions}
\label{sec:Ablation Study}
\subsection{Case Study}

\begin{figure}
\centering     
\subfigure[]{\label{fig:case_11}
                     \includegraphics[width=80mm]{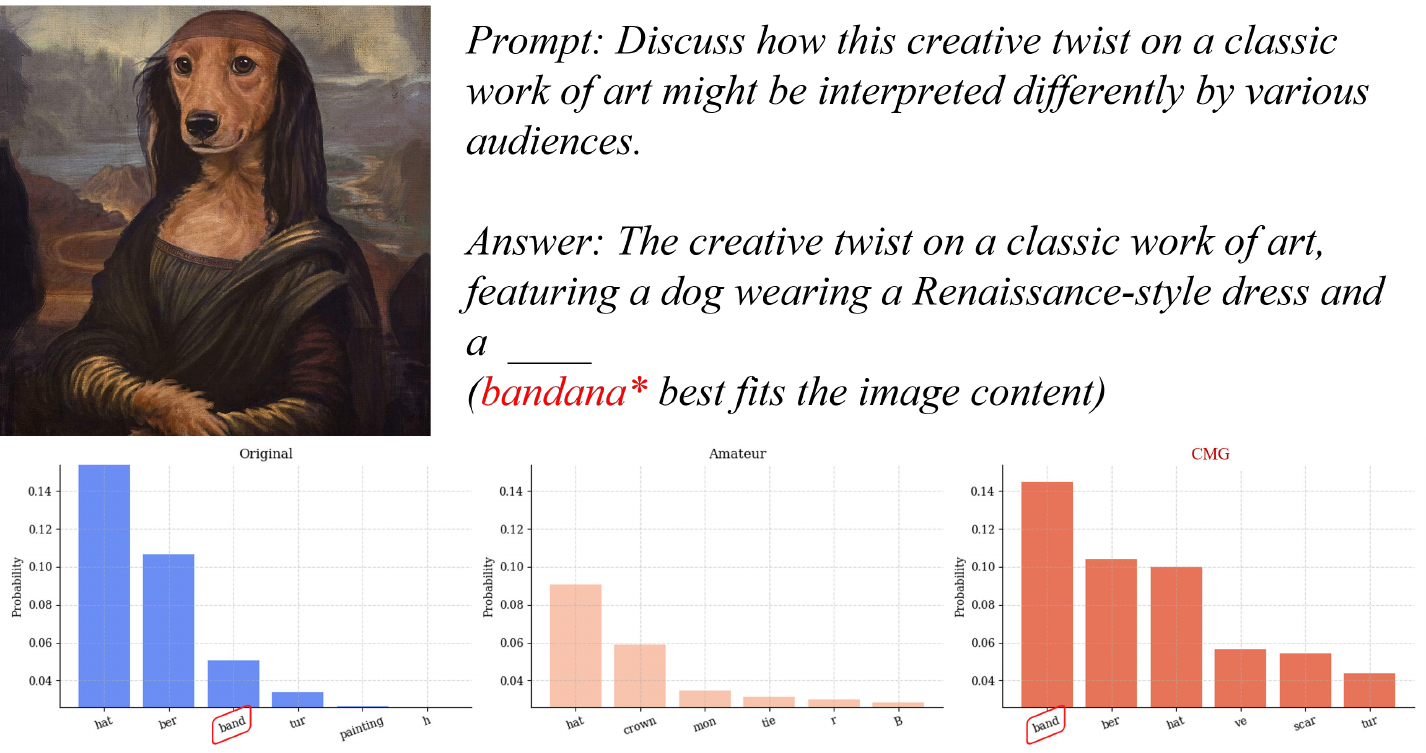}}
\subfigure[]{\label{fig:case_12}
                     \includegraphics[width=80mm]{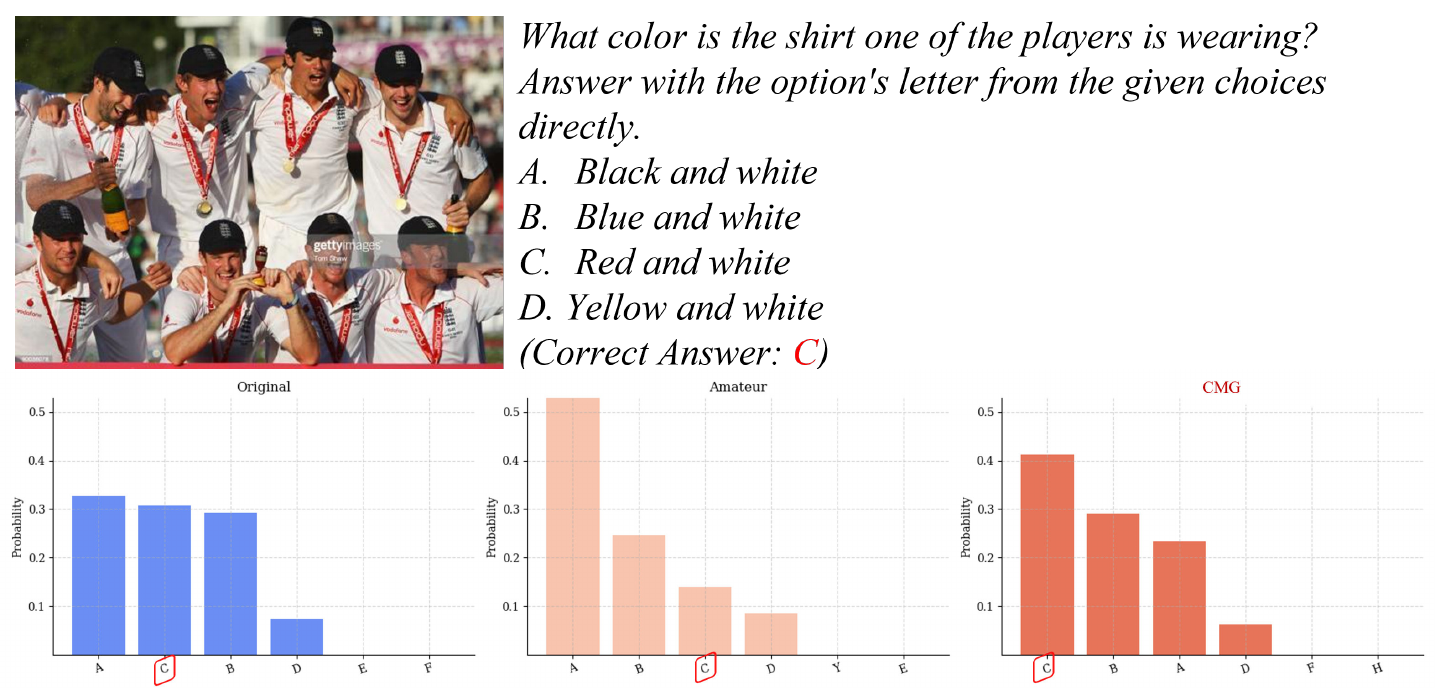}}
\caption{\textbf{Case Study.} \textcolor{red}{Red} boxes circle the correct options.}
\label{fig:Cases}
\end{figure}


\cref{fig:case_11} showcases a painting comprehension task, highlighting the necessity for precise instruction adherence, accurate image interpretation, and pre-trained knowledge. The original model mistakenly associates ``hat" with characters' attire due to skewed co-occurrence biases, a misjudgment exacerbated by the attention-masked model. CMG effectively rectifies this, diminishing the ``hat" confidence and accurately identifying ``bandana" as the correct choice.

In \cref{fig:case_12}, a query about players' T-shirt colors presents a challenge, with black caps potentially misleading VLMs. The original model incorrectly favors option A, failing to discern pertinent visual details. CMG intervention adjusts the PMI ratio, elevating option C's confidence and steering the model towards the accurate response.



\subsection{Unrestrained Amplification of Language Bias in Amateur Models}

\begin{table}
  \centering
  \caption{Results on HallusionBench for Qwen-2-VL-2B. `None*' refers to the ablated model that removes all vision-related attention mechanisms. `Noise*' refers to the ablated model that replaces images with random noise. `Text-only*' denotes the ablated model that converts image inputs into textual captions, substituting the original image with its description.}
\scalebox{0.9} {
  \begin{tabular}{@{}c|c|c|c}
    \toprule
    Method & fAcc & qAcc & aAcc \\
    \midrule
    None* & 12.14 & 16.70 & 53.84  \\
    Noise* & 29.48 & 31.21 & 59.93 \\
    Text-only* & 29.77 & 32.75 & 61.09 \\
    Ours & \textbf{30.06} & \textbf{33.41} & \textbf{61.41} \\
    \bottomrule
  \end{tabular}
}

  \label{tab:MME_noIMG}
\end{table}


In Table~\ref{tab:MME_noIMG}, language bias is amplified by directly removing inter-visual attention and cross-modal attention, but this approach fails to mitigate hallucinations in VLMs; instead, it degrades performance. While removing inter-visual attention and cross-modal attention from inputs, as outlined in \cref{equ:first}, appears to be a straightforward and mathematically consistent solution, our experiments reveal that this method is ineffective. We also explore an alternative amateur model by transforming images into captions to replace the original visual input. This ``Text-only*" method is expected to amplify language bias even compared to CMG, as it introduces accumulated bias through image captioning. However, this approach is also invalid. These findings indicate that the amateur model cannot be constructed arbitrarily; it must occupy a position that is weaker within the original distribution but cannot too weak to answer questions.



\section{Conclusion and Limitation}
\label{sec:Conclusion}

This paper delves into the role of language bias in inducing hallucinations within Vision-Language Models (VLMs) and introduces \emph{Cross-Modal Guidance (CMG)}, an innovative inference strategy designed to counteract such biases. CMG enriches visual context by contrasting outputs from the original model against those from a modified version with disrupted attention maps. Extensive experimentation across various benchmarks has demonstrated CMG's efficacy in bolstering VLM performance.

Despite its advantages, CMG is not without its challenges. It necessitates careful selection of hyper-parameters, like the mask ratio linked to $n_0$ in Eq.~\ref{eqn:optim_st}, suggesting a need for tailored adjustments across different scenarios. Optimal results currently require dynamic hyper-parameter tuning, a complexity we aim to explore further in subsequent research.




\newpage
{
    \small
    \bibliographystyle{ACM-Reference-Format}
    \bibliography{main}


\begin{thebibliography}{36}


\ifx \showCODEN    \undefined \def \showCODEN     #1{\unskip}     \fi
\ifx \showDOI      \undefined \def \showDOI       #1{#1}\fi
\ifx \showISBNx    \undefined \def \showISBNx     #1{\unskip}     \fi
\ifx \showISBNxiii \undefined \def \showISBNxiii  #1{\unskip}     \fi
\ifx \showISSN     \undefined \def \showISSN      #1{\unskip}     \fi
\ifx \showLCCN     \undefined \def \showLCCN      #1{\unskip}     \fi
\ifx \shownote     \undefined \def \shownote      #1{#1}          \fi
\ifx \showarticletitle \undefined \def \showarticletitle #1{#1}   \fi
\ifx \showURL      \undefined \def \showURL       {\relax}        \fi
\providecommand\bibfield[2]{#2}
\providecommand\bibinfo[2]{#2}
\providecommand\natexlab[1]{#1}
\providecommand\showeprint[2][]{arXiv:#2}

\bibitem[Bai et~al\mbox{.}(2023a)]%
        {qwen-lm}
\bibfield{author}{\bibinfo{person}{Jinze Bai}, \bibinfo{person}{Shuai Bai}, \bibinfo{person}{Yunfei Chu}, \bibinfo{person}{Zeyu Cui}, \bibinfo{person}{Kai Dang}, \bibinfo{person}{Xiaodong Deng}, \bibinfo{person}{Yang Fan}, \bibinfo{person}{Wenbin Ge}, \bibinfo{person}{Yu Han}, \bibinfo{person}{Fei Huang}, {et~al\mbox{.}}} \bibinfo{year}{2023}\natexlab{a}.
\newblock \showarticletitle{Qwen technical report}.
\newblock \bibinfo{journal}{\emph{arXiv:2309.16609}} (\bibinfo{year}{2023}).
\newblock


\bibitem[Bai et~al\mbox{.}(2023b)]%
        {Qwen-VL}
\bibfield{author}{\bibinfo{person}{Jinze Bai}, \bibinfo{person}{Shuai Bai}, \bibinfo{person}{Shusheng Yang}, \bibinfo{person}{Shijie Wang}, \bibinfo{person}{Sinan Tan}, \bibinfo{person}{Peng Wang}, \bibinfo{person}{Junyang Lin}, \bibinfo{person}{Chang Zhou}, {and} \bibinfo{person}{Jingren Zhou}.} \bibinfo{year}{2023}\natexlab{b}.
\newblock \showarticletitle{Qwen-VL: A Frontier Large Vision-Language Model with Versatile Abilities}.
\newblock \bibinfo{journal}{\emph{arXiv preprint arXiv:2308.12966}} (\bibinfo{year}{2023}).
\newblock


\bibitem[Chen et~al\mbox{.}(2023a)]%
        {chen2023sharegpt4v}
\bibfield{author}{\bibinfo{person}{Lin Chen}, \bibinfo{person}{Jisong Li}, \bibinfo{person}{Xiaoyi Dong}, \bibinfo{person}{Pan Zhang}, \bibinfo{person}{Conghui He}, \bibinfo{person}{Jiaqi Wang}, \bibinfo{person}{Feng Zhao}, {and} \bibinfo{person}{Dahua Lin}.} \bibinfo{year}{2023}\natexlab{a}.
\newblock \showarticletitle{Sharegpt4v: Improving large multi-modal models with better captions}.
\newblock \bibinfo{journal}{\emph{arXiv:2311.12793}} (\bibinfo{year}{2023}).
\newblock


\bibitem[Chen et~al\mbox{.}(2024)]%
        {chen2024streamliningredundantlayerscompress}
\bibfield{author}{\bibinfo{person}{Xiaodong Chen}, \bibinfo{person}{Yuxuan Hu}, \bibinfo{person}{Jing Zhang}, \bibinfo{person}{Yanling Wang}, \bibinfo{person}{Cuiping Li}, {and} \bibinfo{person}{Hong Chen}.} \bibinfo{year}{2024}\natexlab{}.
\newblock \bibinfo{title}{Streamlining Redundant Layers to Compress Large Language Models}.
\newblock
\newblock
\showeprint[arxiv]{2403.19135}~[cs.CL]
\urldef\tempurl%
\url{https://arxiv.org/abs/2403.19135}
\showURL{%
\tempurl}


\bibitem[Chen et~al\mbox{.}(2020)]%
        {chen2020uniter}
\bibfield{author}{\bibinfo{person}{Yen-Chun Chen}, \bibinfo{person}{Linjie Li}, \bibinfo{person}{Licheng Yu}, \bibinfo{person}{Ahmed El~Kholy}, \bibinfo{person}{Faisal Ahmed}, \bibinfo{person}{Zhe Gan}, \bibinfo{person}{Yu Cheng}, {and} \bibinfo{person}{Jingjing Liu}.} \bibinfo{year}{2020}\natexlab{}.
\newblock \showarticletitle{Uniter: Universal image-text representation learning}. In \bibinfo{booktitle}{\emph{ECCV}}.
\newblock


\bibitem[Chen et~al\mbox{.}(2023b)]%
        {chen2023mitigating}
\bibfield{author}{\bibinfo{person}{Zhiyang Chen}, \bibinfo{person}{Yousong Zhu}, \bibinfo{person}{Yufei Zhan}, \bibinfo{person}{Zhaowen Li}, \bibinfo{person}{Chaoyang Zhao}, \bibinfo{person}{Jinqiao Wang}, {and} \bibinfo{person}{Ming Tang}.} \bibinfo{year}{2023}\natexlab{b}.
\newblock \showarticletitle{Mitigating hallucination in visual language models with visual supervision}.
\newblock \bibinfo{journal}{\emph{arXiv preprint arXiv:2311.16479}} (\bibinfo{year}{2023}).
\newblock


\bibitem[Cho et~al\mbox{.}(2021)]%
        {cho2021unifying}
\bibfield{author}{\bibinfo{person}{Jaemin Cho}, \bibinfo{person}{Jie Lei}, \bibinfo{person}{Hao Tan}, {and} \bibinfo{person}{Mohit Bansal}.} \bibinfo{year}{2021}\natexlab{}.
\newblock \showarticletitle{Unifying vision-and-language tasks via text generation}. In \bibinfo{booktitle}{\emph{ICML}}.
\newblock


\bibitem[Dai et~al\mbox{.}(2023)]%
        {dai2023instructblipgeneralpurposevisionlanguagemodels}
\bibfield{author}{\bibinfo{person}{Wenliang Dai}, \bibinfo{person}{Junnan Li}, \bibinfo{person}{Dongxu Li}, \bibinfo{person}{Anthony Meng~Huat Tiong}, \bibinfo{person}{Junqi Zhao}, \bibinfo{person}{Weisheng Wang}, \bibinfo{person}{Boyang Li}, \bibinfo{person}{Pascale Fung}, {and} \bibinfo{person}{Steven Hoi}.} \bibinfo{year}{2023}\natexlab{}.
\newblock \bibinfo{title}{InstructBLIP: Towards General-purpose Vision-Language Models with Instruction Tuning}.
\newblock
\newblock
\showeprint[arxiv]{2305.06500}~[cs.CV]
\urldef\tempurl%
\url{https://arxiv.org/abs/2305.06500}
\showURL{%
\tempurl}


\bibitem[Driess et~al\mbox{.}(2023)]%
        {palm-e}
\bibfield{author}{\bibinfo{person}{Danny Driess}, \bibinfo{person}{Fei Xia}, \bibinfo{person}{Mehdi~SM Sajjadi}, \bibinfo{person}{Corey Lynch}, \bibinfo{person}{Aakanksha Chowdhery}, \bibinfo{person}{Brian Ichter}, \bibinfo{person}{Ayzaan Wahid}, \bibinfo{person}{Jonathan Tompson}, \bibinfo{person}{Quan Vuong}, \bibinfo{person}{Tianhe Yu}, {et~al\mbox{.}}} \bibinfo{year}{2023}\natexlab{}.
\newblock \showarticletitle{Palm-e: An embodied multimodal language model}.
\newblock \bibinfo{journal}{\emph{arXiv:2303.03378}} (\bibinfo{year}{2023}).
\newblock


\bibitem[Fu et~al\mbox{.}(2024)]%
        {mme}
\bibfield{author}{\bibinfo{person}{Chaoyou Fu}, \bibinfo{person}{Peixian Chen}, \bibinfo{person}{Yunhang Shen}, \bibinfo{person}{Yulei Qin}, \bibinfo{person}{Mengdan Zhang}, \bibinfo{person}{Xu Lin}, \bibinfo{person}{Jinrui Yang}, \bibinfo{person}{Xiawu Zheng}, \bibinfo{person}{Ke Li}, \bibinfo{person}{Xing Sun}, \bibinfo{person}{Yunsheng Wu}, {and} \bibinfo{person}{Rongrong Ji}.} \bibinfo{year}{2024}\natexlab{}.
\newblock \showarticletitle{MME: A Comprehensive Evaluation Benchmark for Multimodal Large Language Models}.
\newblock \bibinfo{journal}{\emph{arxiv:2306.13394}} (\bibinfo{year}{2024}).
\newblock


\bibitem[Gao et~al\mbox{.}(2023)]%
        {llama-adapter-v2}
\bibfield{author}{\bibinfo{person}{Peng Gao}, \bibinfo{person}{Jiaming Han}, \bibinfo{person}{Renrui Zhang}, \bibinfo{person}{Ziyi Lin}, \bibinfo{person}{Shijie Geng}, \bibinfo{person}{Aojun Zhou}, \bibinfo{person}{Wei Zhang}, \bibinfo{person}{Pan Lu}, \bibinfo{person}{Conghui He}, \bibinfo{person}{Xiangyu Yue}, {et~al\mbox{.}}} \bibinfo{year}{2023}\natexlab{}.
\newblock \showarticletitle{Llama-adapter v2: Parameter-efficient visual instruction model}.
\newblock \bibinfo{journal}{\emph{arXiv:2304.15010}} (\bibinfo{year}{2023}).
\newblock


\bibitem[Gong et~al\mbox{.}(2023)]%
        {gong2023multimodalgpt}
\bibfield{author}{\bibinfo{person}{Tao Gong}, \bibinfo{person}{Chengqi Lyu}, \bibinfo{person}{Shilong Zhang}, \bibinfo{person}{Yudong Wang}, \bibinfo{person}{Miao Zheng}, \bibinfo{person}{Qian Zhao}, \bibinfo{person}{Kuikun Liu}, \bibinfo{person}{Wenwei Zhang}, \bibinfo{person}{Ping Luo}, {and} \bibinfo{person}{Kai Chen}.} \bibinfo{year}{2023}\natexlab{}.
\newblock \showarticletitle{MultiModal-GPT: A Vision and Language Model for Dialogue with Humans}.
\newblock \bibinfo{journal}{\emph{arXiv preprint arXiv:2305.04790}} (\bibinfo{year}{2023}).
\newblock


\bibitem[Guan et~al\mbox{.}(2023)]%
        {guan2023hallusionbench}
\bibfield{author}{\bibinfo{person}{Tianrui Guan}, \bibinfo{person}{Fuxiao Liu}, \bibinfo{person}{Xiyang Wu}, \bibinfo{person}{Ruiqi Xian}, \bibinfo{person}{Zongxia Li}, \bibinfo{person}{Xiaoyu Liu}, \bibinfo{person}{Xijun Wang}, \bibinfo{person}{Lichang Chen}, \bibinfo{person}{Furong Huang}, \bibinfo{person}{Yaser Yacoob}, {et~al\mbox{.}}} \bibinfo{year}{2023}\natexlab{}.
\newblock \showarticletitle{HallusionBench: An Advanced Diagnostic Suite for Entangled Language Hallucination and Visual Illusion in Large Vision-Language Models}.
\newblock \bibinfo{journal}{\emph{arXiv preprint arXiv:2310.14566}} (\bibinfo{year}{2023}).
\newblock


\bibitem[Gunjal et~al\mbox{.}(2024)]%
        {gunjal2024detectingpreventinghallucinationslarge}
\bibfield{author}{\bibinfo{person}{Anisha Gunjal}, \bibinfo{person}{Jihan Yin}, {and} \bibinfo{person}{Erhan Bas}.} \bibinfo{year}{2024}\natexlab{}.
\newblock \bibinfo{title}{Detecting and Preventing Hallucinations in Large Vision Language Models}.
\newblock
\newblock
\showeprint[arxiv]{2308.06394}~[cs.CV]
\urldef\tempurl%
\url{https://arxiv.org/abs/2308.06394}
\showURL{%
\tempurl}


\bibitem[Holtzman et~al\mbox{.}(2020)]%
        {holtzman2020curiouscaseneuraltext}
\bibfield{author}{\bibinfo{person}{Ari Holtzman}, \bibinfo{person}{Jan Buys}, \bibinfo{person}{Li Du}, \bibinfo{person}{Maxwell Forbes}, {and} \bibinfo{person}{Yejin Choi}.} \bibinfo{year}{2020}\natexlab{}.
\newblock \bibinfo{title}{The Curious Case of Neural Text Degeneration}.
\newblock
\newblock
\showeprint[arxiv]{1904.09751}~[cs.CL]
\urldef\tempurl%
\url{https://arxiv.org/abs/1904.09751}
\showURL{%
\tempurl}


\bibitem[Leng et~al\mbox{.}(2024)]%
        {leng2024mitigating}
\bibfield{author}{\bibinfo{person}{Sicong Leng}, \bibinfo{person}{Hang Zhang}, \bibinfo{person}{Guanzheng Chen}, \bibinfo{person}{Xin Li}, \bibinfo{person}{Shijian Lu}, \bibinfo{person}{Chunyan Miao}, {and} \bibinfo{person}{Lidong Bing}.} \bibinfo{year}{2024}\natexlab{}.
\newblock \showarticletitle{Mitigating object hallucinations in large vision-language models through visual contrastive decoding}. In \bibinfo{booktitle}{\emph{Proceedings of the IEEE/CVF Conference on Computer Vision and Pattern Recognition}}. \bibinfo{pages}{13872--13882}.
\newblock


\bibitem[Li et~al\mbox{.}(2021)]%
        {li2021align}
\bibfield{author}{\bibinfo{person}{Junnan Li}, \bibinfo{person}{Ramprasaath Selvaraju}, \bibinfo{person}{Akhilesh Gotmare}, \bibinfo{person}{Shafiq Joty}, \bibinfo{person}{Caiming Xiong}, {and} \bibinfo{person}{Steven Chu~Hong Hoi}.} \bibinfo{year}{2021}\natexlab{}.
\newblock \showarticletitle{Align before fuse: Vision and language representation learning with momentum distillation}.
\newblock \bibinfo{journal}{\emph{NeurIPS}} (\bibinfo{year}{2021}).
\newblock


\bibitem[Li et~al\mbox{.}(2023a)]%
        {li2023contrastivedecodingopenendedtext}
\bibfield{author}{\bibinfo{person}{Xiang~Lisa Li}, \bibinfo{person}{Ari Holtzman}, \bibinfo{person}{Daniel Fried}, \bibinfo{person}{Percy Liang}, \bibinfo{person}{Jason Eisner}, \bibinfo{person}{Tatsunori Hashimoto}, \bibinfo{person}{Luke Zettlemoyer}, {and} \bibinfo{person}{Mike Lewis}.} \bibinfo{year}{2023}\natexlab{a}.
\newblock \bibinfo{title}{Contrastive Decoding: Open-ended Text Generation as Optimization}.
\newblock
\newblock
\showeprint[arxiv]{2210.15097}~[cs.CL]
\urldef\tempurl%
\url{https://arxiv.org/abs/2210.15097}
\showURL{%
\tempurl}


\bibitem[Li et~al\mbox{.}(2023b)]%
        {Li-pope}
\bibfield{author}{\bibinfo{person}{Yifan Li}, \bibinfo{person}{Du Yifan}, \bibinfo{person}{Zhou Kun}, \bibinfo{person}{Wang Jinpeng}, \bibinfo{person}{Wayne}, \bibinfo{person}{Zhao Xin}, {and} \bibinfo{person}{Wen Ji-Rong}.} \bibinfo{year}{2023}\natexlab{b}.
\newblock \showarticletitle{Evaluating Object Hallucination in Large Vision-Language Models}. In \bibinfo{booktitle}{\emph{The 2023 Conference on Empirical Methods in Natural Language Processing}}.
\newblock
\urldef\tempurl%
\url{https://openreview.net/forum?id=xozJw0kZXF}
\showURL{%
\tempurl}


\bibitem[Liu et~al\mbox{.}(2023a)]%
        {llava}
\bibfield{author}{\bibinfo{person}{Haotian Liu}, \bibinfo{person}{Chunyuan Li}, \bibinfo{person}{Qingyang Wu}, {and} \bibinfo{person}{Yong~Jae Lee}.} \bibinfo{year}{2023}\natexlab{a}.
\newblock \showarticletitle{Visual instruction tuning}.
\newblock \bibinfo{journal}{\emph{arXiv:2304.08485}} (\bibinfo{year}{2023}).
\newblock


\bibitem[Liu et~al\mbox{.}(2023b)]%
        {liu2023dejavucontextualsparsity}
\bibfield{author}{\bibinfo{person}{Zichang Liu}, \bibinfo{person}{Jue Wang}, \bibinfo{person}{Tri Dao}, \bibinfo{person}{Tianyi Zhou}, \bibinfo{person}{Binhang Yuan}, \bibinfo{person}{Zhao Song}, \bibinfo{person}{Anshumali Shrivastava}, \bibinfo{person}{Ce Zhang}, \bibinfo{person}{Yuandong Tian}, \bibinfo{person}{Christopher Re}, {and} \bibinfo{person}{Beidi Chen}.} \bibinfo{year}{2023}\natexlab{b}.
\newblock \bibinfo{title}{Deja Vu: Contextual Sparsity for Efficient LLMs at Inference Time}.
\newblock
\newblock
\showeprint[arxiv]{2310.17157}~[cs.LG]
\urldef\tempurl%
\url{https://arxiv.org/abs/2310.17157}
\showURL{%
\tempurl}


\bibitem[Luo et~al\mbox{.}(2024)]%
        {luo2024monointernvlpushingboundariesmonolithic}
\bibfield{author}{\bibinfo{person}{Gen Luo}, \bibinfo{person}{Xue Yang}, \bibinfo{person}{Wenhan Dou}, \bibinfo{person}{Zhaokai Wang}, \bibinfo{person}{Jifeng Dai}, \bibinfo{person}{Yu Qiao}, {and} \bibinfo{person}{Xizhou Zhu}.} \bibinfo{year}{2024}\natexlab{}.
\newblock \bibinfo{title}{Mono-InternVL: Pushing the Boundaries of Monolithic Multimodal Large Language Models with Endogenous Visual Pre-training}.
\newblock
\newblock
\showeprint[arxiv]{2410.08202}~[cs.CV]
\urldef\tempurl%
\url{https://arxiv.org/abs/2410.08202}
\showURL{%
\tempurl}


\bibitem[Maaz et~al\mbox{.}(2023)]%
        {maaz2023videochatgpt}
\bibfield{author}{\bibinfo{person}{Muhammad Maaz}, \bibinfo{person}{Hanoona Rasheed}, \bibinfo{person}{Salman Khan}, {and} \bibinfo{person}{Fahad~Shahbaz Khan}.} \bibinfo{year}{2023}\natexlab{}.
\newblock \showarticletitle{Video-ChatGPT: Towards Detailed Video Understanding via Large Vision and Language Models}.
\newblock \bibinfo{journal}{\emph{arXiv preprint arXiv:2306.05424}} (\bibinfo{year}{2023}).
\newblock


\bibitem[OpenAI(2023)]%
        {openai2023gpt4}
\bibfield{author}{\bibinfo{person}{OpenAI}.} \bibinfo{year}{2023}\natexlab{}.
\newblock \showarticletitle{GPT-4 technical report}.
\newblock \bibinfo{journal}{\emph{arXiv:2303.08774}} (\bibinfo{year}{2023}).
\newblock


\bibitem[Ouyang et~al\mbox{.}(2022)]%
        {ouyang2022traininglanguagemodelsfollow}
\bibfield{author}{\bibinfo{person}{Long Ouyang}, \bibinfo{person}{Jeff Wu}, \bibinfo{person}{Xu Jiang}, \bibinfo{person}{Diogo Almeida}, \bibinfo{person}{Carroll~L. Wainwright}, \bibinfo{person}{Pamela Mishkin}, \bibinfo{person}{Chong Zhang}, \bibinfo{person}{Sandhini Agarwal}, \bibinfo{person}{Katarina Slama}, \bibinfo{person}{Alex Ray}, \bibinfo{person}{John Schulman}, \bibinfo{person}{Jacob Hilton}, \bibinfo{person}{Fraser Kelton}, \bibinfo{person}{Luke Miller}, \bibinfo{person}{Maddie Simens}, \bibinfo{person}{Amanda Askell}, \bibinfo{person}{Peter Welinder}, \bibinfo{person}{Paul Christiano}, \bibinfo{person}{Jan Leike}, {and} \bibinfo{person}{Ryan Lowe}.} \bibinfo{year}{2022}\natexlab{}.
\newblock \bibinfo{title}{Training language models to follow instructions with human feedback}.
\newblock
\newblock
\showeprint[arxiv]{2203.02155}~[cs.CL]
\urldef\tempurl%
\url{https://arxiv.org/abs/2203.02155}
\showURL{%
\tempurl}


\bibitem[Park et~al\mbox{.}(2024)]%
        {park2024convis}
\bibfield{author}{\bibinfo{person}{Yeji Park}, \bibinfo{person}{Deokyeong Lee}, \bibinfo{person}{Junsuk Choe}, {and} \bibinfo{person}{Buru Chang}.} \bibinfo{year}{2024}\natexlab{}.
\newblock \showarticletitle{ConVis: Contrastive Decoding with Hallucination Visualization for Mitigating Hallucinations in Multimodal Large Language Models}.
\newblock \bibinfo{journal}{\emph{arXiv preprint arXiv:2408.13906}} (\bibinfo{year}{2024}).
\newblock


\bibitem[Radford et~al\mbox{.}(2021)]%
        {clip}
\bibfield{author}{\bibinfo{person}{Alec Radford}, \bibinfo{person}{Jong~Wook Kim}, \bibinfo{person}{Chris Hallacy}, \bibinfo{person}{Aditya Ramesh}, \bibinfo{person}{Gabriel Goh}, \bibinfo{person}{Sandhini Agarwal}, \bibinfo{person}{Girish Sastry}, \bibinfo{person}{Amanda Askell}, \bibinfo{person}{Pamela Mishkin}, \bibinfo{person}{Jack Clark}, {et~al\mbox{.}}} \bibinfo{year}{2021}\natexlab{}.
\newblock \showarticletitle{Learning transferable visual models from natural language supervision}. In \bibinfo{booktitle}{\emph{ICML}}.
\newblock


\bibitem[Shi et~al\mbox{.}(2023)]%
        {shi2023trustingevidencehallucinatecontextaware}
\bibfield{author}{\bibinfo{person}{Weijia Shi}, \bibinfo{person}{Xiaochuang Han}, \bibinfo{person}{Mike Lewis}, \bibinfo{person}{Yulia Tsvetkov}, \bibinfo{person}{Luke Zettlemoyer}, {and} \bibinfo{person}{Scott~Wen tau Yih}.} \bibinfo{year}{2023}\natexlab{}.
\newblock \bibinfo{title}{Trusting Your Evidence: Hallucinate Less with Context-aware Decoding}.
\newblock
\newblock
\showeprint[arxiv]{2305.14739}~[cs.CL]
\urldef\tempurl%
\url{https://arxiv.org/abs/2305.14739}
\showURL{%
\tempurl}


\bibitem[Su et~al\mbox{.}(2023)]%
        {pandagpt}
\bibfield{author}{\bibinfo{person}{Yixuan Su}, \bibinfo{person}{Tian Lan}, \bibinfo{person}{Huayang Li}, \bibinfo{person}{Jialu Xu}, \bibinfo{person}{Yan Wang}, {and} \bibinfo{person}{Deng Cai}.} \bibinfo{year}{2023}\natexlab{}.
\newblock \showarticletitle{PandaGPT: One Model To Instruction-Follow Them All}.
\newblock \bibinfo{journal}{\emph{arXiv:2305.16355}} (\bibinfo{year}{2023}).
\newblock


\bibitem[Wang et~al\mbox{.}(2024)]%
        {Qwen2VL}
\bibfield{author}{\bibinfo{person}{Peng Wang}, \bibinfo{person}{Shuai Bai}, \bibinfo{person}{Sinan Tan}, \bibinfo{person}{Shijie Wang}, \bibinfo{person}{Zhihao Fan}, \bibinfo{person}{Jinze Bai}, \bibinfo{person}{Keqin Chen}, \bibinfo{person}{Xuejing Liu}, \bibinfo{person}{Jialin Wang}, \bibinfo{person}{Wenbin Ge}, \bibinfo{person}{Yang Fan}, \bibinfo{person}{Kai Dang}, \bibinfo{person}{Mengfei Du}, \bibinfo{person}{Xuancheng Ren}, \bibinfo{person}{Rui Men}, \bibinfo{person}{Dayiheng Liu}, \bibinfo{person}{Chang Zhou}, \bibinfo{person}{Jingren Zhou}, {and} \bibinfo{person}{Junyang Lin}.} \bibinfo{year}{2024}\natexlab{}.
\newblock \showarticletitle{Qwen2-VL: Enhancing Vision-Language Model's Perception of the World at Any Resolution}.
\newblock \bibinfo{journal}{\emph{arXiv preprint arXiv:2409.12191}} (\bibinfo{year}{2024}).
\newblock


\bibitem[Wang et~al\mbox{.}(2021)]%
        {wang2021simvlm}
\bibfield{author}{\bibinfo{person}{Zirui Wang}, \bibinfo{person}{Jiahui Yu}, \bibinfo{person}{Adams~Wei Yu}, \bibinfo{person}{Zihang Dai}, \bibinfo{person}{Yulia Tsvetkov}, {and} \bibinfo{person}{Yuan Cao}.} \bibinfo{year}{2021}\natexlab{}.
\newblock \showarticletitle{Simvlm: Simple visual language model pretraining with weak supervision}.
\newblock \bibinfo{journal}{\emph{arXiv:2108.10904}} (\bibinfo{year}{2021}).
\newblock


\bibitem[Yue et~al\mbox{.}(2024)]%
        {yue2024mmmu}
\bibfield{author}{\bibinfo{person}{Xiang Yue}, \bibinfo{person}{Yuansheng Ni}, \bibinfo{person}{Kai Zhang}, \bibinfo{person}{Tianyu Zheng}, \bibinfo{person}{Ruoqi Liu}, \bibinfo{person}{Ge Zhang}, \bibinfo{person}{Samuel Stevens}, \bibinfo{person}{Dongfu Jiang}, \bibinfo{person}{Weiming Ren}, \bibinfo{person}{Yuxuan Sun}, {et~al\mbox{.}}} \bibinfo{year}{2024}\natexlab{}.
\newblock \showarticletitle{Mmmu: A massive multi-discipline multimodal understanding and reasoning benchmark for expert agi}. In \bibinfo{booktitle}{\emph{Proceedings of the IEEE/CVF Conference on Computer Vision and Pattern Recognition}}. \bibinfo{pages}{9556--9567}.
\newblock


\bibitem[Zhan et~al\mbox{.}(2025)]%
        {zhan2025griffon}
\bibfield{author}{\bibinfo{person}{Yufei Zhan}, \bibinfo{person}{Yousong Zhu}, \bibinfo{person}{Zhiyang Chen}, \bibinfo{person}{Fan Yang}, \bibinfo{person}{Ming Tang}, {and} \bibinfo{person}{Jinqiao Wang}.} \bibinfo{year}{2025}\natexlab{}.
\newblock \showarticletitle{Griffon: Spelling out all object locations at any granularity with large language models}. In \bibinfo{booktitle}{\emph{European Conference on Computer Vision}}. Springer, \bibinfo{pages}{405--422}.
\newblock


\bibitem[Zhang et~al\mbox{.}(2023b)]%
        {zhang2023videollama}
\bibfield{author}{\bibinfo{person}{Hang Zhang}, \bibinfo{person}{Xin Li}, {and} \bibinfo{person}{Lidong Bing}.} \bibinfo{year}{2023}\natexlab{b}.
\newblock \showarticletitle{Video-LLaMA: An Instruction-tuned Audio-Visual Language Model for Video Understanding}.
\newblock \bibinfo{journal}{\emph{arXiv preprint arXiv:2306.02858}} (\bibinfo{year}{2023}).
\newblock


\bibitem[Zhang et~al\mbox{.}(2023a)]%
        {llama-adapter}
\bibfield{author}{\bibinfo{person}{Renrui Zhang}, \bibinfo{person}{Jiaming Han}, \bibinfo{person}{Aojun Zhou}, \bibinfo{person}{Xiangfei Hu}, \bibinfo{person}{Shilin Yan}, \bibinfo{person}{Pan Lu}, \bibinfo{person}{Hongsheng Li}, \bibinfo{person}{Peng Gao}, {and} \bibinfo{person}{Yu Qiao}.} \bibinfo{year}{2023}\natexlab{a}.
\newblock \showarticletitle{Llama-adapter: Efficient fine-tuning of language models with zero-init attention}.
\newblock \bibinfo{journal}{\emph{arXiv:2303.16199}} (\bibinfo{year}{2023}).
\newblock


\bibitem[Zhu et~al\mbox{.}(2023)]%
        {minigpt-4}
\bibfield{author}{\bibinfo{person}{Deyao Zhu}, \bibinfo{person}{Jun Chen}, \bibinfo{person}{Xiaoqian Shen}, \bibinfo{person}{Xiang Li}, {and} \bibinfo{person}{Mohamed Elhoseiny}.} \bibinfo{year}{2023}\natexlab{}.
\newblock \showarticletitle{Minigpt-4: Enhancing vision-language understanding with advanced large language models}.
\newblock \bibinfo{journal}{\emph{arXiv:2304.10592}} (\bibinfo{year}{2023}).
\newblock


\end{thebibliography}
}


\end{document}